\documentclass[journal]{IEEEtran}
\usepackage{graphicx}
\usepackage{subfigure}
\usepackage[hidelinks]{hyperref}

\usepackage{caption}
\usepackage{float}
\usepackage{amsfonts,amssymb}
\usepackage{booktabs}
\let\orgautoref\autoref

\renewcommand{\autoref}[1]{\def\equationautorefname{Eq.}
\def\figureautorefname{Fig.}\def\subfigureautorefname{Fig.}\orgautoref{#1}}
\usepackage{multirow}
\usepackage{cite}
\usepackage{amsmath}
\usepackage{algorithm}
\usepackage{algorithmic}
\usepackage{array}
\newcommand{\PreserveBackslash}[1]{\let\temp=\\#1\let\\=\temp}
\newcolumntype{C}[1]{>{\PreserveBackslash\centering}p{#1}}
\newcolumntype{R}[1]{>{\PreserveBackslash\raggedleft}p{#1}}
\newcolumntype{L}[1]{>{\PreserveBackslash\raggedright}p{#1}}

\usepackage{fixltx2e}
\begin{document}
%
\title{Multistage Model for Robust Face Alignment Using Deep Neural Networks}
%
%
%

\author{Huabin~Wang$^1$, Rui~Cheng$^{1}$, Jian~Zhou$^1$, Liang~Tao$^1$, and Hon~Keung~Kwan$^{2}$
	\\$^1$MOE Key Laboratory of Intelligent Computing and Signal Processing, School of Computer Science and Technology, Anhui University, Hefei, Anhui 230601 China.
	\\$^2$Department of Electrical and Computer Engineering, University of Windsor, Windsor, Ontario N9B 3P4 Canada.
	\\\{wanghuabin, jzhou, taoliang\}@ahu.edu.cn, chengrui@stu.ahu.edu.cn, kwan1@uwindsor.ca
}

\maketitle

\begin{abstract}
An ability to generalize unconstrained conditions such as severe occlusions and large pose variations remains a challenging goal to achieve in face alignment. In this paper, a multistage model based on deep neural networks is proposed which takes advantage of spatial transformer networks, hourglass networks and exemplar-based shape constraints. First, a spatial transformer - generative adversarial network which consists of convolutional layers and residual units is utilized to solve the initialization issues caused by face detectors, such as rotation and scale variations, to obtain improved face bounding boxes for face alignment. Then, stacked hourglass network is employed to obtain preliminary locations of landmarks as well as their corresponding scores. In addition, an exemplar-based shape dictionary is designed to determine landmarks with low scores based on those with high scores. By incorporating face shape constraints, misaligned landmarks caused by occlusions or cluttered backgrounds can be considerably improved. Extensive experiments based on challenging benchmark datasets are performed to demonstrate the superior performance of the proposed method over other state-of-the-art methods.
\end{abstract}


%
\IEEEpeerreviewmaketitle

\begin{figure*}[htpb]
	\centering
	\includegraphics[width=\linewidth]{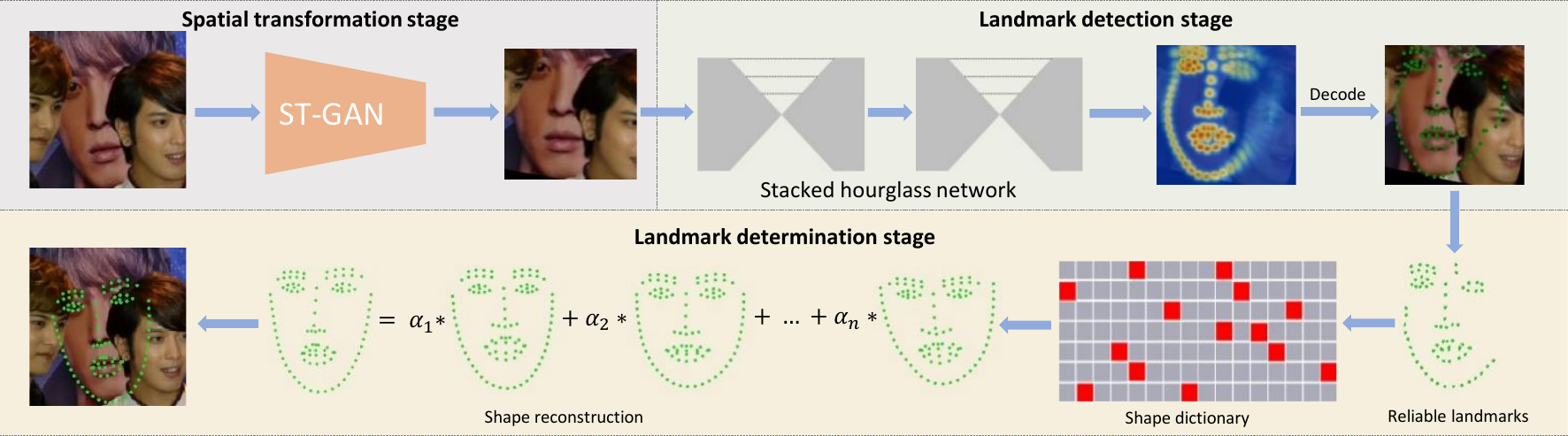}
	\caption{Overview of the proposed multistage model (MSM). First, spatial transformer - generative adversarial network (ST-GAN) normalizes a face to a canonical state. Second, stacked hourglass network is used to obtain score maps, which determine the position and confidence score of each landmark. Finally, landmarks with high scores are used to search for similar shapes from the shape dictionary; and landmarks with low scores are determined by a weighted combination of all score maps using reconstruction coefficients, $\alpha_i$.}
	\label{fig:pipeline}
\end{figure*}
%

\section{Introduction}
%
%
%
%
\IEEEPARstart{F}{ace} alignment (or facial landmark detection) aims to locate a set of predefined human facial landmarks, such as the corners of the eyes, the eyebrows, and the tip of the nose for high-level vision tasks, such as face recognition\cite{liu2018joint}, expression recognition \cite{shan2009facial}, facial animation\cite{cao2014displaced} and 3D face modelling \cite{jourabloo2016large}. Although considerable progress has been made, face alignment is still challenging due to large-view face variations, lighting conditions, complex expressions and partial occlusions.

Recently, progresses have been made by convolutional neural networks (CNNs) in semantic segmentation\cite{long2015fully} and in human pose estimation and face alignment based on heatmap regression\cite{newell2016stacked}. The hourglass network\cite{newell2016stacked} offers a method for human pose estimation. The model utilizes repeated down-sampled and up-sampled modules to extract features across multiple scales. The hourglass network has been introduced to face alignment task and achieved efficient performance. However, existing methods are still inefficient in modelling face structural priors, the performance of these methods degrades severely when face images suffer from heavy occlusion, and this problem is challenging  to address since occlusion is common and diverse in reality.

Several typical face alignment models have attempted to address faces under partial occlusions. Robust cascaded pose regression (RCPR)\cite{burgos2013robust} is the first method that simultaneously detects landmarks and estimates occlusions. In this method, the face is divided into a $3\times3$ grid for each regression stage, and only one non-occluded face region is used to predict the location of the landmarks. The work in \cite{wu2015robust} proposed a unified framework that combines landmark localization and visibility estimation, which focuses more on landmarks with high visibility probabilities and iteratively updates landmark locations and landmark visibility probabilities. Xing \textit{et al.}\cite{xing2018towards} considered the regression procedure as a sparse coding problem by learning two dictionaries: one is the face appearance dictionary, the other is the face shape dictionary. With two relational dictionaries, the occluded face appearance is restored, and the influence of the occluded landmarks is suppressed. Liu \textit{et al.}\cite{liu2017adaptive} utilized shape-indexed appearance to estimate the occlusion level of each landmark, and the face shape is reconstructed by similar shapes from the exemplar-based shape dictionary. Although these methods have shown superior performance in detecting occluded landmarks, they still suffer from poor scalability and robustness. The first limitation is the lack of large-scale ground truth occlusion annotation for natural images. The task of providing occlusion annotation is often time consuming, involving a considerable amount of tedious manual work. Additionally, due to the inherent complex variations in human facial appearance in unconstrained environments, it is difficult to recover the occluded appearance using face appearance dictionary.

Another challenge is the initialization issue of face images derived from face detectors, which has drawn little attention in previous studies. The pre-processing step of face alignment is to crop face rectangles through a face detector. However, due to severe occlusion or blur, the face detector may not produce an appropriate face rectangle. As Ren \textit{et al.} noted in \cite{ren2016face}, if the initial images have different scale and rotation variations, the performance of many face alignment methods would be severely degraded. It will be useful if an algorithm could produce canonical face poses with the same scales and centre shifts. The work of \cite{lv2017deep} proposed a deep regression framework with two-stage reinitialization to address the problems of face image initialization and landmark detection. In this model, the spatial transformer networks (STNs) is embedded as subnets at each stage. However, due to its complex architecture and end-to-end learning strategy, the STN is hard to be supervised during training, or worse yet, has a negative impact on the performance of final coordinates regression. In\cite{yang2017stacked}, a simple regression network is employed to detect several facial key points, and then performed Procrustes analysis with the mean shape to obtain affine transformation parameters, further removing the rigid transformation. However, under severe occlusion conditions, even the state-of-the-art algorithms may fail to localize landmarks correctly, to make matters worse, the inaccurate locations of landmarks lead to the inaccurate prediction of affine transformation parameters.

In this work\footnote{This work is built on top of\cite{yan2018score} with four major contributions as listed at the end of Section I.}, a multistage model (MSM) is proposed to address the problem of face image initialization and to facilitate the robustness of face alignment under occlusion. The MSM consists of three parts: a spatial transformer - generative adversarial network (ST-GAN), a two-stage hourglass network and an exemplar-based shape dictionary.\autoref{fig:pipeline} gives an overview of MSM. First, ST-GAN produces better initial facial images by removing rigid transformations from translation, scale and rotation. In contrast to the original STN\cite{jaderberg2015spatial}, the idea of adversarial learning\cite{goodfellow2014generative} is introduced to enhance the accuracy of spatial transformation. STN is considered a generator; then, a discriminator is designed to distinguish whether the pose of the generated facial image is canonical. After facial image initialization, canonical facial images are fed to the hourglass network. The output of the hourglass network consists of a set of score maps, and each score map determines the primary position and reliability score for each landmark. The reliability score is used to measure the quality of the localization. The key innovation of MSM is that landmarks with high scores are utilized to refine the landmarks with low scores. Specifically, due to partial occlusion, the occluded landmarks cannot be located precisely, and the visible landmark can be predicted precisely. As shown in\autoref{fig:pipeline}, the scores of visible landmarks are high in the heatmap and the landmarks under occlusion have lower scores than the visible landmarks. Thus, reliable landmarks with high scores can help to refine the occluded landmarks with low scores. Finally, an exemplar-based shape dictionary is introduced to search for the most similar shapes and reconstruct the face shape based on the landmarks with high scores.

In summary, we make the following contributions to the face alignment task:
\begin{enumerate}
	\item A spatial transformer - generative adversarial network is proposed to produce promising initial face images for face alignment.
	\item Based on the intensity of the heatmaps obtained by a two stage hourglass network, a scoring scheme is designed to measure the quality of predicted landmarks locations, which can estimate the occlusion level of each landmark and distinguish the aligned landmarks from misaligned landmarks.
	\item An exemplar-based shape dictionary is employed to impose geometric constraints. The landmarks with high scores are used to search similar shapes from dictionary, and the landmarks with low scores are refined by shape reconstruction using similar shapes.
	\item Experiment results on several benchmark datasets (300-W, COFW and WFLW) show that the proposed multistage model outperforms most recent face alignment methods, especially for faces with difficult scenarios such as large pose, lighting and occlusion, etc.
\end{enumerate}

\section{Related Work}
In this section, we first review the development of face alignment, and then briefly review STNs.

\subsection{Face Alignment}
Face alignment methods can be generally classified into three categories: discriminative fitting, cascaded shape regression, and deep learning.

Since facial shape and facial appearance are deformable structured objects, methods based on discriminative fitting typically model facial structures by learning shape and appearance variation models. According to the difference in facial  representations, these methods can be divided into two categories: one is the holistic-based representation, such as active appearance model (AAM)\cite{cootes2001active}, the other is part-based representation, such as active shape model (ASM)\cite{cootes1995active}, constrained local model (CLM)\cite{cristinacce2006feature}, Gauss-Newton deformable part model (GN-DPM)\cite{tzimiropoulos2014gauss}. These methods typically require an iterative process to find the optimal parameter configuration for a given face, thus it is time-consuming and prone to fall into local minima. Moreover, due to the limited capacity of parametric models, such methods are sensitive to occlusion and large pose variation.

Methods based on cascaded shape regression were popular in face alignment before the advent of deep learning. These approaches are based on a multistage framework, and each stage refines the position of predicted landmarks in a coarse-to-fine manner. Specifically, a weak regressor is utilized in each stage to model the relation between the image feature and the shape increment. Cootes \textit{et al.}\cite{cootes2012robust} proposed an efficient method that combines random forest regression and a statistical shape model. The supervised descent method (SDM)\cite{xiong2013supervised} focuses on solving the optimization problem of the least squares method. Ren \textit{et al.}\cite{ren2016face} proposed learning local binary features around local patches using random forest regression, which was faster than existing methods. In \cite{fan2018explicit}, a projective invariant is designed for modelling the intrinsic structure of human faces and combined it with cascade regression methods. The regression-based approach mentioned above employs the handcrafted feature descriptors (e.g., SIFT\cite{xiong2013supervised}, HoG\cite{yan2013learn}, or random forest/fern descriptors\cite{ren2016face}) to extract facial texture information. It is clear that conventional cascaded regression methods have yielded drastic improvements based on standard benchmarks such as 300-W\cite{sagonas2013300}. However, most of these methods are sensitive to initialized shapes, due to the limitations of handcrafted features. 

Recently, CNNs have made a series of breakthroughs in many visual analysis tasks such as image classification\cite{he2016deep}, semantic segmentation\cite{long2015fully}, and human pose estimation\cite{newell2016stacked}. The application of CNNs greatly boosts the performance of face alignment. CNN-based methods can be generally classified into two categories: coordinate regression methods\cite{sun2013deep,zhang2016learning,xiao2016robust} and heatmap regression methods\cite{kowalski2017deep,bulat2017binarized,bulat2017far,deng2019joint,wu2018look,valle2018deeply}. The difference between the two categories is that the former directly regresses landmark coordinates with a network, and the latter first learns a mapping function from image to likelihood heatmaps, and chooses the location with the highest response value in the heatmap as the predicted location. Sun \textit{et al.}\cite{sun2013deep} first introduced CNNs to the face alignment field, and cascaded three CNNs to detect facial landmarks in a multistage manner. The method in \cite{zhang2016learning} jointly learns landmark localization and correlated recognition tasks, such as facial attributes and expressions. Xiao \textit{et al.}\cite{xiao2016robust} proposed a framework that leverages the advantages of CNNs and recurrent neural networks (RNNs). The feature extraction stage is replaced with a CNN, and the fitting stage is replaced with an RNN. Weng \textit{et al.}\cite{weng2016learning} proposed an exemplar-based cascaded auto-encoder network for real-time face alignment. These coordinate regression methods can directly detect the coordinates of landmarks and do not require post-processing operations. However, since coordinate regression methods are predicted landmarks from dense layers that contain high-level semantic information but lack the details of the facial texture, result in limitations in real-world scenarios, such as occlusion, large poses, and other uncontrolled conditions. Kowalski \textit{et al.}\cite{kowalski2017deep} first introduced the idea of heatmaps to cascaded CNNs. They generated heatmaps based on the predicted coordinates of the previous stage, and then combined the original image as an input for the next stage. In\cite{bulat2017far}, a binary hourglass network with a multi-scale feature fusion residual module is developed to boost performance for 2D and 3D face alignment. Deng \textit{et al.}\cite{deng2019joint} employed affine transformation to remove rotation and scale variations in facial images and then detected landmarks through hourglass networks. In\cite{wu2018look}, the concept of boundary heatmap is introduced as a facial geometry. Valle \textit{et al.}\cite{valle2018deeply} combined a CNN and ensemble of regression trees (ERT) to enhance computational efficiency. Although heatmap regression methods represented by hourglass networks show excellent performance, there are still many limitations for hourglass networks to model the geometric structure of the human face.
\subsection{Spatial Transformer Network}
CNNs achieve excellent performance in local feature representation. However, CNNs still lack the ability to be spatially invariant to the input image. Jaderberg \textit{et al.}\cite{jaderberg2015spatial} first presented STN that explicitly learns invariance to translation, scale and rotation. Benefiting from STN, they achieved state-of-the-art performance in several image classification tasks, such as MNIST\cite{deng2012mnist} digit classification. STN allows a neural network to learn how to perform spatial transformations on an input image to enhance the geometric invariance of the model. In \cite{chen2016supervised}, an STN was embedded in cascaded CNNs, to jointly learn spatial transformation and landmark localization for face detection. Similarly, the work of\cite{lv2017deep} embedded an STN as a subnet to obtain an improved initial image for face landmark localization. In\cite{lin2018st}, STN is applied to the task of image composition, and an STN is embedded in the generator of the generative adversarial network (GAN) for warping a specific object of a given image and placing it in the scene image. Apparently original STN is robust to handling the spatial transformation of simple objects, such as handwritten digits. Due to the complex variations of faces in uncontrolled conditions, the original STN has difficulty in robustly providing accurate spatial transformations.

\section{Method}
As illustrated in\autoref{fig:pipeline}, MSM consists of three pivotal steps: GAN-based spatial transformation, CNN-based landmark detection and exemplar-based shape reconstruction. In this section, MSM is described in detail.
\subsection{Spatial Transformer - Generative Adversarial Network}
\begin{figure*}[!t]
	\centering
	\includegraphics[width=\linewidth]{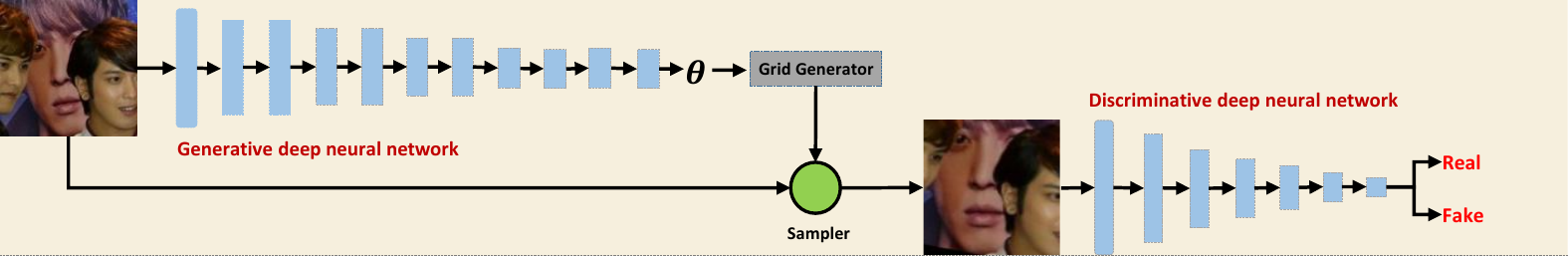}
	\caption{Architecture of spatial transformer - generative adversarial network (ST-GAN). The generative deep neural network (GDNN) is used to generate the transformation matrix $\theta$. The discriminative deep neural network (DDNN) is used to determine whether the generated face image is ``real", which means a canonical face without unnecessary background.}
	\label{fig:stn}
\end{figure*}
Recent studies\cite{ren2016face,lv2017deep} have shown that the pre-processing of face images is critical to face alignment tasks. If the initialized image has a large pose or excessive unnecessary background, the accuracy of landmark localization is greatly reduced. There are two typical methods for facial image pre-processing: one is based on affine transformation, and the other is based on STNs. Affine transformation methods first detect several fiducial key points and then calculate the parameters of affine transformation by Procrustes analysis based on located key points and the key points of the mean face shape. It is obvious that affine transformation methods have the same limitations as the conventional face alignment algorithm, regarding sensitivity to occlusion and blur. STN-based methods explicitly learn image warping without key point detection, which is more flexible and robust than the affine transformation approach. Nonetheless, due to the complexity of the human face in nature, it is challenging to regress accurate transformation parameters using the basic STN model.

To improve the robustness of STN\cite{jaderberg2015spatial} to handling complex face images, adversarial learning is introduced. As shown in\autoref{fig:stn}, the proposed spatial transformer - generative adversarial network (ST-GAN) consists of two parts: a generative deep neural network (GDNN) and a discriminative deep neural network (DDNN). Similar to original STN\cite{jaderberg2015spatial}, the generative deep neural network consists of three main components: a localization network, a grid generator and a sampler. The localization network is realized by a convolutional network consisting of 11 convolutional layers with different strides.
%
The overall configuration of the proposed GDNN and DDNN are listed in\autoref{table:structure of G} and\autoref{table:structure of D}, respectively. The size of input of GDNN is $128\times128$.
Each of the first 9 convolutional layers of the GDNN is of size $3\times3$ with different strides. At the end, a $4\times4$ global average pooling layer and a $1\times1$ convolutional layer are utilized to regress the transformation matrix $\mathcal{\theta}$. For 2D affine transformation, the transformation matrix $\mathcal{\theta}$ is selected to be a 2 by 3 matrix.
\begin{equation}\label{equ:transformation parameter}
\mathcal{\theta}=\left(\begin{array}{ccc}
\theta_{11} & \theta_{12} & \theta_{13}\\
\theta_{21} & \theta_{22} & \theta_{23}\\
\end{array}
\right)
\end{equation}

\begin{table}[!ht]
	\centering
	\setlength{\abovecaptionskip}{2pt}
	\caption{ST-GAN architecture. Configuration refers to size, number of convolutional kernels, and number of strides.}
	\label{table:structure of G}
	\centering
	\scriptsize
	\begin{tabular}{c|c|c|c}
		\hline
		\rule[0.5ex]{0pt}{1.5ex}Layer	&Input size   &Configuration &Output size  \\ \hline
		\rule[0.5ex]{0pt}{1.5ex}Conv1	&$128\times128\times3$  &$3\times3$, 8, stride 2	&$64\times64\times8$  \\ \hline
		\rule[0.5ex]{0pt}{1.5ex}Conv2	&$64\times64\times8$  &$3\times3$, 16, stride 2  &$32\times32\times16$   \\ \hline
		\rule[0.5ex]{0pt}{1.5ex}Conv3	&$32\times32\times16$  &$3\times3$, 16, stride 1  &$32\times32\times16$    \\ \hline
		\rule[0.5ex]{0pt}{1.5ex}Conv4	&$32\times32\times16$  &$3\times3$, 32, stride 2  &$16\times16\times32$   \\ \hline
		\rule[0.5ex]{0pt}{1.5ex}Conv5	&$16\times16\times32$  &$3\times3$, 32, stride 1  &$16\times16\times32$    \\ \hline
		\rule[0.5ex]{0pt}{1.5ex}Conv6	&$16\times16\times32$  &$3\times3$, 16, stride 2  &$8\times8\times64$   \\	\hline
		\rule[0.5ex]{0pt}{1.5ex}Conv7	&$8\times8\times64$  &$3\times3$, 64, stride 1  &$8\times8\times64$    \\ \hline
		\rule[0.5ex]{0pt}{1.5ex}Conv8	&$8\times8\times64$  &$3\times3$, 16, stride 2  &$4\times4\times128$   \\	\hline
		\rule[0.5ex]{0pt}{1.5ex}Conv9	&$4\times4\times128$  &$3\times3$, 128, stride 1  &$4\times4\times128$    \\ \hline
		\rule[0.5ex]{0pt}{1.5ex}Conv10	&$4\times4\times128$  &$4\times4$, 32, stride 1	&$1\times1\times32$  \\ \hline
		\rule[0.5ex]{0pt}{1.5ex}Conv11	&$1\times1\times32$  &$1\times1$, 6, stride 1	&$1\times1\times6$  \\ \hline
	\end{tabular}
\end{table}

\begin{table}[!ht]
	\centering
	\setlength{\abovecaptionskip}{2pt}
	\caption{DDNN architecture. Configuration refers to size, number of convolutional kernels, and number of strides.}
	\label{table:structure of D}
	\centering
	\scriptsize
	\begin{tabular}{c|c|c|c}
		\hline
		\rule[0.5ex]{0pt}{1.5ex}Layer	&Input size   &Configuration	 &Output size  \\ \hline
		\rule[0.5ex]{0pt}{1.5ex}Conv1	&$128\times128\times3$  &$4\times4$, 32, stride 2	&$64\times64\times32$  \\ \hline
		\rule[0.5ex]{0pt}{1.5ex}Conv2	&$64\times64\times32$  &$4\times4$, 64, stride 2	&$32\times32\times64$  \\ \hline
		\rule[0.5ex]{0pt}{1.5ex}Conv3	&$32\times32\times64$  &$4\times4$, 128, stride 2	&$16\times16\times128$  \\ \hline
		\rule[0.5ex]{0pt}{1.5ex}Conv4	&$16\times16\times128$  &$4\times4$, 256, stride 2	&$8\times8\times256$  \\ \hline
		\rule[0.5ex]{0pt}{1.5ex}Conv5	&$8\times8\times256$  &$4\times4$, 512, stride 2	&$4\times4\times512$  \\ \hline
		\rule[0.5ex]{0pt}{1.5ex}Conv6	&$4\times4\times512$  &$4\times4$, 1024, stride 2	&$2\times2\times1024$  \\ \hline
		\rule[0.5ex]{0pt}{1.5ex}Conv7	&$2\times2\times1024$  &$2\times2$, 2, stride 1	&$1\times1\times2$  \\ \hline
	\end{tabular}
\end{table}

The grid generator generates a grid $R=\{g_i\}, g_i=[x_i, y_i]$ in the input image corresponding to each pixel $i$ from the output image. The sampler uses the transformation matrix $\mathcal{\theta}$ and applies it to the input image. Specifically, assuming $(x_i^s,y_i^s)$ are the source coordinates of the $i$-th of the input image and $ (x_i^t, y_i^t) $ are the target coordinates of the $i$-th of the output image, the transformation procedure is defined as follows.
\begin{equation}
\left( \begin{array}{c}\label{equ:transformation procedure}
			x_i^s\\
			y_i^s\\
			\end{array}
			\right)=\mathcal{\theta}(g)
			=
			\left( \begin{array}{ccc}
			\theta_{11} & \theta_{12} & \theta_{13}\\
			\theta_{21} & \theta_{22} & \theta_{23}\\
			\end{array}
			\right)\left( \begin{array}{c}
			x_i^t\\
			y_i^t\\
			1
			\end{array}
			\right)
\end{equation}

\begin{figure*}[!t]
	\centering 
	\setlength{\abovecaptionskip}{5pt}
	\scalebox{1}{\includegraphics[width=1.025\linewidth]{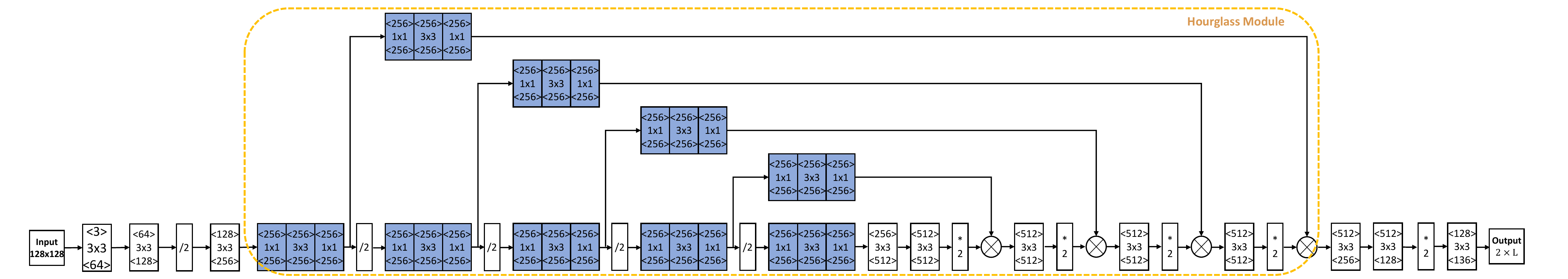}}
	\caption{Architecture of single hourglass network. Each set of 3 rectangular boxes represents one residual unit. The numbers in the angle brackets at the top and bottom of the each blue rectangle indicate the number of channels of the input feature map and output feature map, respectively. ``$/2$" and ``$*2$" denote a max pooling layer and a deconvolutional layer, respectively. Finally, the output is a $2\times L$ vector, $L$ denotes the total number of landmarks in a face image.}
	\label{fig:hourglass}
\end{figure*}

Similar to\cite{lv2017deep}, supervised learning is applied to train affine transformation parameters. As shown in\autoref{table:structure of D}, the size of the input of DDNN is $128\times128$, and the output is a scalar representing the possibilities. Each of the first 6 convolutional layers is of size $4\times4$ with stride 2, the convolutional layer 7 is of size $2\times2$ with stride 1. The loss function of discriminator DDNN is defined as follows (for simplicity, GDNN is denoted as $G$, DDNN is denoted as $D$):
\begin{equation}\label{equ:loss_D}
\mathcal{L}_D=\mathbb{E}[\log{D(I_{real})}]+\mathbb{E}[\log{(1-D(G(I_{fake})))}]
\end{equation}
where $I_{real}$ refers to real sample which is the ground truth image without rotation, scale and unnecessary background. $I_{fake}$ refers to noise sample which is a designed facial image with rotation, scale and unnecessary background. $\mathbb{E}$ represents the expectation. The discriminator learns to predict the ground truth facial image as one while predicting the generated facial image as zero. With DDNN, the adversarial loss can be defined as follows:
\begin{equation}\label{equ:loss_A}
\mathcal{L}_A = \mathbb{E}[\log(1-D(G(I_{fake})))]
\end{equation}
The loss function of generator $G$ is defined as
\begin{equation}\label{equ:loss_G}
\mathcal{L}_G= a ||\hat{\mathcal{\theta}}-\mathcal{\theta^*}||+ b \mathcal{L}_A
\end{equation}
where $\hat{\mathcal{\theta}}$ is the parameter regressed by GDNN and $\mathcal{\theta^*}$ is the ground truth transformation parameter. The hyper-parameters $a$ and $b$ are used to balance different losses. Thus, GDNN is optimized to fool discriminator DDNN by regressing more accurate parameter that will improve the learning of the spatial transformation.
The final objective function can be expressed as follows.
\begin{equation}\label{equ:loss_whole}
\mathop{\arg}\mathop{\min}_{G}\mathop{\max}_{D} (\mathcal{L}_G + \mathcal{L}_D)
\end{equation}
In this way, the generator $G$ and the discriminator $D$ play a minimax game in which $D$ tries to maximize the probability it correctly classifies the face pose is canonical or not (i.e. real or fake), and $G$ tries to minimize the probability that $D$ will predict its output is fake. The whole training process is summarized in \textbf{Algorithm}\autoref{alg:stn}.

\begin{algorithm}[!t]
	\caption{Training process of ST-GAN.}
	\label{alg:stn}
	\begin{algorithmic}[1]
		\REQUIRE Training images $I_{fake}$, the corresponding ground-truth image $I_{real}$, the generator $G$, the discriminator $D$.\STATE Forward $G$ by $G(I)$, and optimize $G$ according to \autoref{equ:loss_G};\STATE Forward $D$ by $D(I_{real})$ and optimize $D$ by maximizing the first term of $\mathcal{L}_D$ defined in \autoref{equ:loss_D};
		\STATE Forward $D$ by $D(I_{fake})$ and optimize $D$ by maximizing the second term of $\mathcal{L}_D$ defined in \autoref{equ:loss_D};
		\STATE Optimize $G$ by \autoref{equ:loss_whole};
		\STATE Go back to \textbf{Step} 1 until the accuracy of the validation set stop increasing;
		\STATE return $G$.
	\end{algorithmic}
\end{algorithm}

\subsection{CNN-based preliminary landmark detection}

Exemplar-based sparse constraints require a set of reliable landmarks to converge. Thus, the objective of the preliminary stage is to precisely locate visible landmarks. Deep convolutional neural network is an effective method for detecting visible landmarks. Stacked hourglass network\cite{newell2016stacked}, which is a repeated encoder and decoder architecture, has proven to have some distinct advantages: 1) It is a simple, minimally designed network with the capability of capturing information at different scales; 2) In a symmetrical topology, two feature maps with the same resolution are connected by skip connections to better maintain low-level information; 3) There is a loss function for intermediate supervision at the end of each hourglass module; 4) It can produce pixel-wise predictions of the same resolution as the input image. Recently, many work adopted four or eight hourglass modules as network backbone, but such strategy are computationally expensive for real-time applications.

\begin{figure}[!t]
	\centering 
	\includegraphics[width=0.43\linewidth]{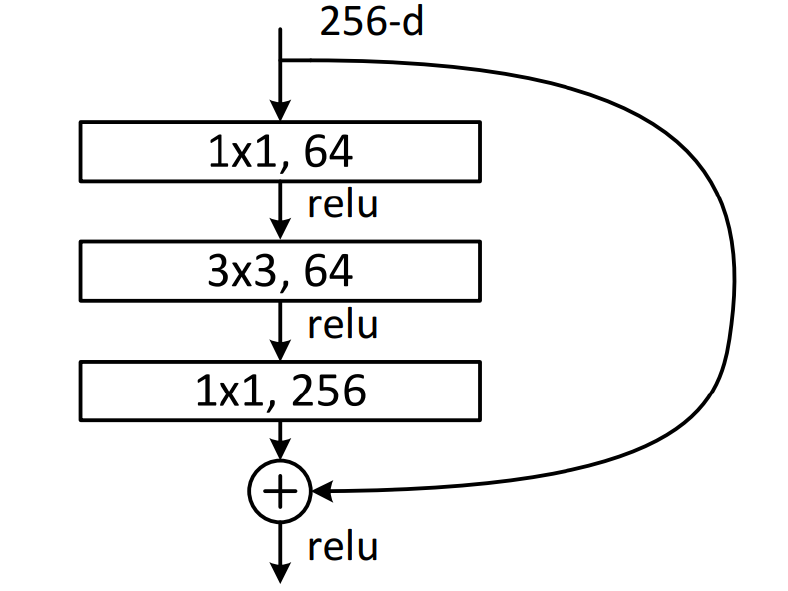}
	\caption{Structure of a residual unit.}
	\label{fig:residual_unit}
\end{figure}

To achieve a good trade-off between performance and efficiency, a network based on two hourglass modules is designed. Residual unit\cite{he2016deep} are used as the building blocks in the hourglass network,\autoref{fig:residual_unit} gives the detail of a 3-layer residual unit. A residual block can be expressed as follows:
\begin{equation}
x_{n+1}= x_{n}+F(x_{n}, W_{n})
\end{equation}
Where $x_{n+1}$ and $x_{n}$ are the output and input feature maps of the $n$-th block, $W_{n}$ denotes the weights of convolutional layers. $F$ consists of batch normalization, ReLU is used for non linearity function, two $1\times1$ convolutional layers and a $3\times3$ convolutional layer, with an $1\times1$ skip convolutional layer are used to match different channels of input and output feature maps. Stacked residual units can increase feature channels and extract high-level discriminative features. First, we give an overview of the network architecture. As shown in\autoref{fig:hourglass}, the input of the network is a face image normalized by the previous ST-GAN with a spatial resolution of $128\times128$, followed by two $3\times3$ convolutional layers to increase the number of feature channels and a max pooling layer to decrease the resolution from 128 to 64, through a $3\times3$ convolutional layer and a residual unit, the number of channels is increased to 256. then the feature maps with 256 channels and $64\times64$ resolution are fed to the hourglass module. The hourglass module consists of a four-layer recursive structure, and each level consists of a downsampling layer, residual units, a skip connection layer and a deconvolutional layer. Considering computational costs, $64\times64$ resolution is used in the hourglass module. Unlike the original hourglass module\cite{newell2016stacked} which uses upsampling layer to recover the size of the feature maps, deconvolution\cite{zeiler2010deconvolutional} is introduced to replace upsampling layers to better maintain spatial semantic information. Batch normalization is performed before all convolutional layers to accelerate convergence except for the first convolutional layer with $3\times3$ kernels. ReLU is used as an activation function.

\begin{figure}[!t]
	\centering 
	\includegraphics[width=\linewidth]{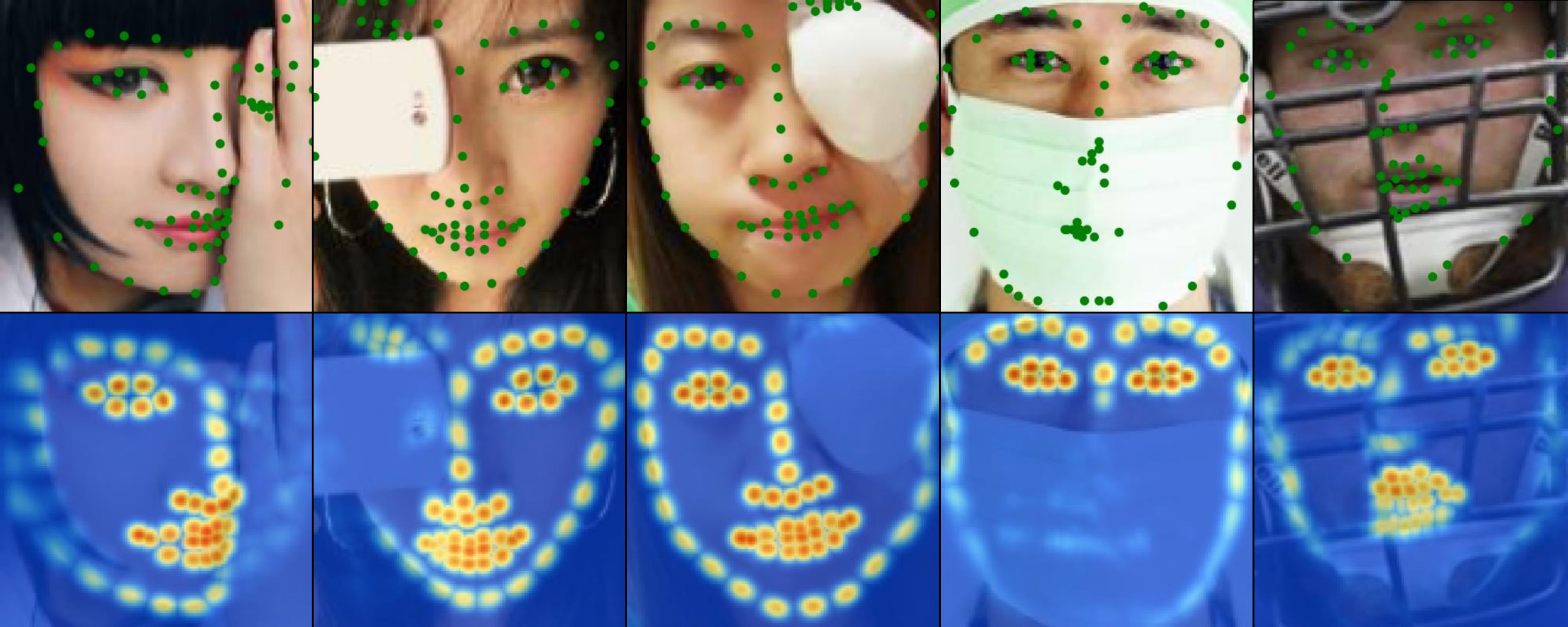}
	\caption{Example outputs obtained by two-stage hourglass network. The first row shows detected landmark locations. The second row shows the corresponding heatmaps. Note that the occluded landmarks cannot be precisely located in most cases. The non-occluded landmarks in heatmaps have higher intensity values than the occluded ones.}
	\label{fig:heatmap_on_visible_and_occlusion}
\end{figure}

For an image $I$, this network is trained to obtain $L$ heatmaps $H(I)$, where $L$ is the total number of landmarks for each face. The location of each predicted landmark is decoded from corresponding heatmap by taking the location with the maximum value as follows:
\begin{equation}
\mathrm{c}(l)=\arg\max H^l(I)
\end{equation}
where $l$ is the index of the landmark and the corresponding heatmap. $\mathrm{c}(l)$ gives the coordinate of the $l$-th landmark. Some examples output by this network are shown in\autoref{fig:heatmap_on_visible_and_occlusion}. Note that visible landmarks can be precisely located; however, these results may not have a biological human facial shape since occluded landmarks were not detected. In addition, the response heatmaps of visible landmarks are more focused than those of occluded landmarks. It is challenging to decode corrected positions from scattered heatmaps, which is a limitation of the heatmap regression-based method. 

To review the definition of a heatmap. During training, a ground truth heatmap for one landmark is created by putting a Gaussian peak at the ground truth location of a landmark, and the intensity decreases with the distance to the closest landmark. Motivated by a recent study\cite{liu2017adaptive} that used shape-indexed appearance to estimate the occlusion level of each landmark, the intensity of the heatmap is employed to estimate location quality and further distinguish reliable landmarks and missing landmarks. In detail, each landmark is weighted based on the corresponding intensity values in the heatmaps. Thus, more reliable landmarks with strong local information are assigned high weights. The landmarks under occlusion are assigned low weights. The process of assigning weight can be expressed by the following equation:
\begin{equation}\label{equ:score assignment}
w_l = \dfrac{\sum_{k=X_l-r}^{X_l+r}\sum_{t=Y_l-r}^{Y_l+r}score_l(k,t)}{(2\times r+1)^2}
\end{equation}
where $score_l(k,t)$ is the value of coordinate $(k, t)$ in the $l$-th heatmap, $r$ determines the size of the rectangle used to calculate the score. The coordinate $(X_l, Y_l)$ gives the predicted location of the $l$-th landmark. Based on the assigned weight, the predicted landmarks can be classified into two categories: reliable landmarks and misaligned landmarks. The coordinate and weights of reliable landmarks act as initial information for the following shape refinement stage.

\begin{figure*}[!t]
	\centering
	\includegraphics[width=\linewidth]{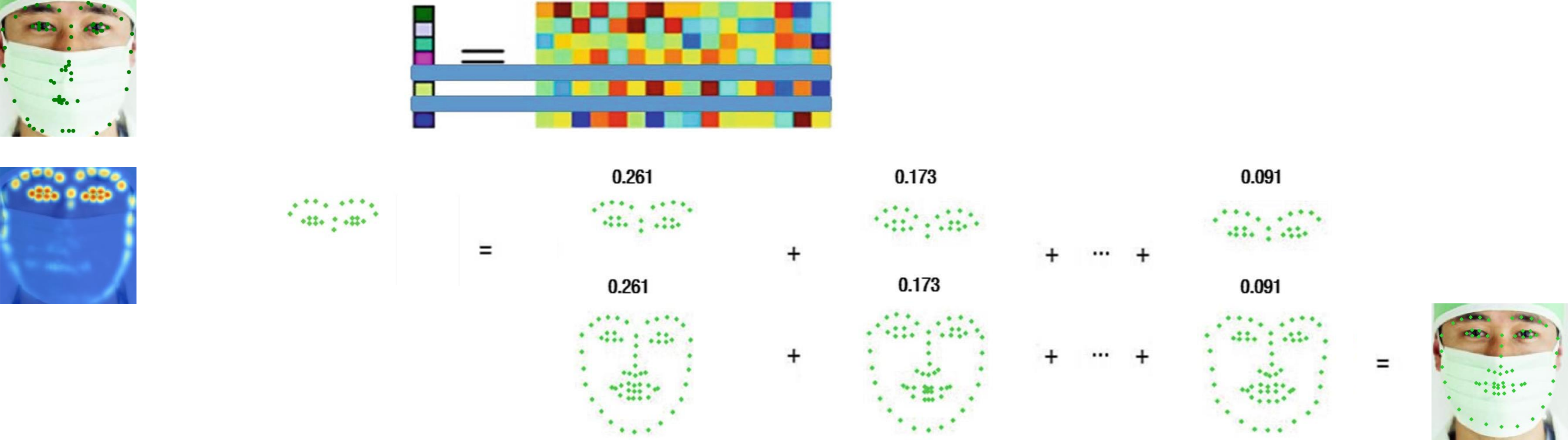}
	\caption{Face shape reconstruction based on nearest exemplar shapes. The reconstruction target is a partial face shape which consists only reliable landmarks.}
	\label{fig:shape reconstruct}
\end{figure*}

\subsection{Exemplar-based Shape Reconstruction}
Deep convolutional neural networks have a strong capacity for local feature representation, thus the visible landmarks can be effectively located through the first two stages. However, a large number of parameters can easily lead to network overfitting, especially for limited training samples. In addition, CNNs still lack the ability to model the geometric structure of the human face, resulting in sensitivity to occlusion. In contrast, human vision is capable of predicting face shapes by utilizing geometric constraints. Motivated by this ability, these misaligned landmarks can be refined by similar face shapes in the training samples, and this approach is feasible and simple. To this end, following \cite{liu2016dual,liu2017adaptive}, sparse shape constraints are incorporated to correct the misaligned landmarks. The sparse shape model is a popular method of imposing shape priors, it can refine the gross error and maintains shape detail at the same time. This feature allows the model to be perfectly integrated with CNNs. The objective of the sparse shape model can be formulated as follows:
\begin{equation}
\arg\min\left\|S-D_s\alpha\right\|_2 + \lambda\left\|\alpha\right\|_2
\end{equation}
where $S$ is a $2L\times1$ vector with $L$ landmark coordinates of the predicted normalized shape. $D_s$ is an $N\times2L$ matrix, that is a shape dictionary with a sample size of $N$. $\alpha$ is the shape reconstruction coefficient, and $\lambda$ is the regularization parameter. As Liu \textit{et al.} noted in\cite{liu2017adaptive}, the traditional sparse shape model treats all landmarks equally, causing the error from corrupted landmarks spread to other aligned landmarks, and harms the convergence of the model. In other words, incorrect reconstruction targets lead the sparse shape constraint to produce incorrect shapes. Different from\cite{liu2017adaptive}, only the accurately aligned landmarks which were assigned high weights are used to search for similar shapes from a dictionary. As shown in\autoref{fig:shape reconstruct}, this part of the facial shape, which consists of only reliable landmarks, is our reconstruction target.

After the first two stages, the preliminary coordinates and weight of each landmark can be determined. Then a threshold $T$ is set to distinguish reliable landmarks and misaligned landmarks, Thus, for each shape $S$ we obtained a binary vector $V$. If the $l$-th component of $V$ is $1$, then the $l$-th landmark is considered reliable. Based on reliable landmarks, the search process can be formulated as follows:
\begin{equation}
\min_{\alpha}\|V^*S-(V^*S\odot V^*D_S)\alpha\|_2^2
\end{equation}
where $V^*=\mathrm{\mathit{diag}}(V)$. The goal of $V^*$ is to force the search process to neglect misaligned landmarks and emphasize landmarks with high weights. $\odot$ indicates searching for the most similar shape in the dictionary. $(V^*S\odot V^*D_S)$ is used to search for the $k$ nearest exemplar shapes of $V^*S$ from the adaptive shape dictionary $V^*D_s$. Then the misaligned part shape can be reconstructed by the $k$ nearest shapes and the reconstruction coefficients can be simply computed by the least squares method. However, searching all training samples is time consuming, especially for a large training set. Furthermore, there are many similar face shapes that are redundant. Thus, K-means algorithm is applied to all training shapes to obtain $N$ representative face shapes, which form a compact shape dictionary $D_S$. Searching from $D_S$ will be more effective. The shape reconstruction procedure is shown in\autoref{fig:shape reconstruct}. The whole process of the proposed multistage model is summarized in \textbf{Algorithm}\autoref{alg:shape reconstruction}.
\begin{algorithm}[h]
	\caption{Multistage Model}
	\label{alg:shape reconstruction}
	\begin{algorithmic}[1]
		\REQUIRE Face image $I$, face rectangle $R$, shape dictionary $D_S$, threshold $T$.
		\STATE Crop $I$ according to $R$, get facial part image $I_R$.
		\STATE Feed $I_R$ to ST-GAN, get normalize face image $I_N$ and transform parameter $\theta$.
		\STATE Feed $I_N$ to stacked hourglass network, get preliminary face shape $S$. 
		\STATE Calculate weight $w_i$ by\autoref{equ:score assignment} for each landmark.
		\STATE \textbf{for} $i=1$ to $L$ \textbf{do}
		\STATE \quad $w_i=w_i*\max_{i=1,2,...L}(w_i)$
		\STATE \quad \textbf{if} $w_i>T$ \textbf{then} $V_i=1$
		\STATE \quad \textbf{else} $V_i=0$
		\STATE \quad \textbf{end if}
		\STATE \textbf{end}
		\STATE $V^*=\mathrm{\mathit{diag}}(V)$.
		\STATE Shape reconstruction via \\
		 $\arg\min_{\alpha}\|V^*S-(V^*S\odot V^*D_S)\alpha\|_2^2$.
		\STATE Final shape $S_F=D_S\alpha$.
		\STATE Affine to original resolution by $S_O=\theta^{-1}S_F$.
	\end{algorithmic}
\end{algorithm}




\section{Experiments}
In this section, we conduct extensive experiments and analysis to show the effectiveness of the proposed method. The following paragraphs describe the datasets, implementation details, experimental results and ablation study.
\subsection{Datasets}
Our method is evaluated on several challenging datasets including 300-W, COFW and WFLW.
\subsubsection{300-W\cite{sagonas2013300}} 300-W is currently the most widely used dataset. It was created from four datasets including the AFW\cite{ramanan2012face}, LFPW\cite{belhumeur2013localizing}, HELEN\cite{le2012interactive} and IBUG\cite{sagonas2013300} dataset, each face image
is annotated with 68 landmarks. The training set consists of the AFW, LFPW training set and HELEN training set, resulting in a total of 3148 images. The test set consists of three parts: the common set, challenge set and full set. The common set consists of the LPFW test set and HELEN test set, resulting in a total of 554 images. The challenge set, which is the IBUG dataset, contains 135 images. The full set consists of a common set and challenge set containing 689 images.
\subsubsection{300-W private test set\cite{sagonas2016300}} The 300-W private test set was introduced after the 300-W dataset and was used for the 300-W Challenge benchmark. It consists of 300 indoor images and 300 outdoor images, each image was annotated 68 landmarks using the same annotation scheme as the one of 300-W.
\subsubsection{COFW\cite{burgos2013robust}} The COFW dataset focuses on occlusion in nature. The training set consists of 1345 images, the testing set consists of 507 faces with a wide range of occlusion patterns, and each face is annotated with 29 landmarks. In our experiment we use reannotated version\cite{ghiasi2014occlusion} of the 68 landmarks for comparison to other approaches.
\subsubsection{WFLW\cite{wu2018look}} WFLW is considered the most challenging dataset. It contains 10000 faces (7500 for training and 2500 for testing) with 98 fully manually annotated landmarks and corresponding facial bounding boxes. Compared to the above datasets, WFLW includes rich attribute annotations, such as occlusion, pose, make-up, blur and illumination attribute information. 

\subsection{Evaluation Metrics}
Similar to previous methods, we use the normalized root mean squared error (NRMSE), cumulative errors distribution (CED) curve, area under the curve (AUC) and failure rate to measure the landmark location error.
\begin{equation}\label{NRMSE}
NRMSE=\frac{1}{N}\sum_{i}^{N}\frac{\frac{1}{L}\sum_{j}^{L}|P_{ij}-G_{ij}|_2}{d_i}
\end{equation}
where $N$ is the number of total images, $L$ is the number of total landmarks for a given face, and $P_{ij}$ and $G_{ij}$ denote the predicted and ground truth locations, respectively. $d_i$ is the normalization parameter. The experiment results using different definitions of $d_i$: the distance between the eye centres (“inter-pupils”) and the distance between the outer eye corners (“inter-ocular”).

For the 300-W, 300-W test set and COFW dataset, image with an NRMSE (“inter-ocular”) of 0.08 or greater is considered a failure. For the WFLW dataset, following\cite{wu2018look}, image with an NRMSE (“inter-ocular”) of 0.1 or greater is considered a failure.

\subsection{Implementation Details}
We independently trained three models: ST-GAN, stacked hourglass network and face shape dictionary. For ST-GAN, the faces are cropped by the provided bounding boxes and resized to 128$\times$128 resolution. Data augmentation is applied by random flipping, rotation (between $\pm30^{\circ}$), scaling (between $\pm10\%$) and colour jittering. The network is optimized by Adam stochastic optimization with an initial learning rate of 0.0005 and reduced by half after 400 epochs. In total, 1000 epochs are used in training. The mini batch size is set to 16. The stacked hourglass network was trained following a similar procedure, and the difference is that the input images of the network are cropped by ground truth bounding boxes, training is applied for a total of 300 epochs. The learning rate is reduced to half after 100 epochs. Both networks were implemented in PyTorch.

\begin{figure}[ht!]
	\centering
	\includegraphics[width=0.6\linewidth]{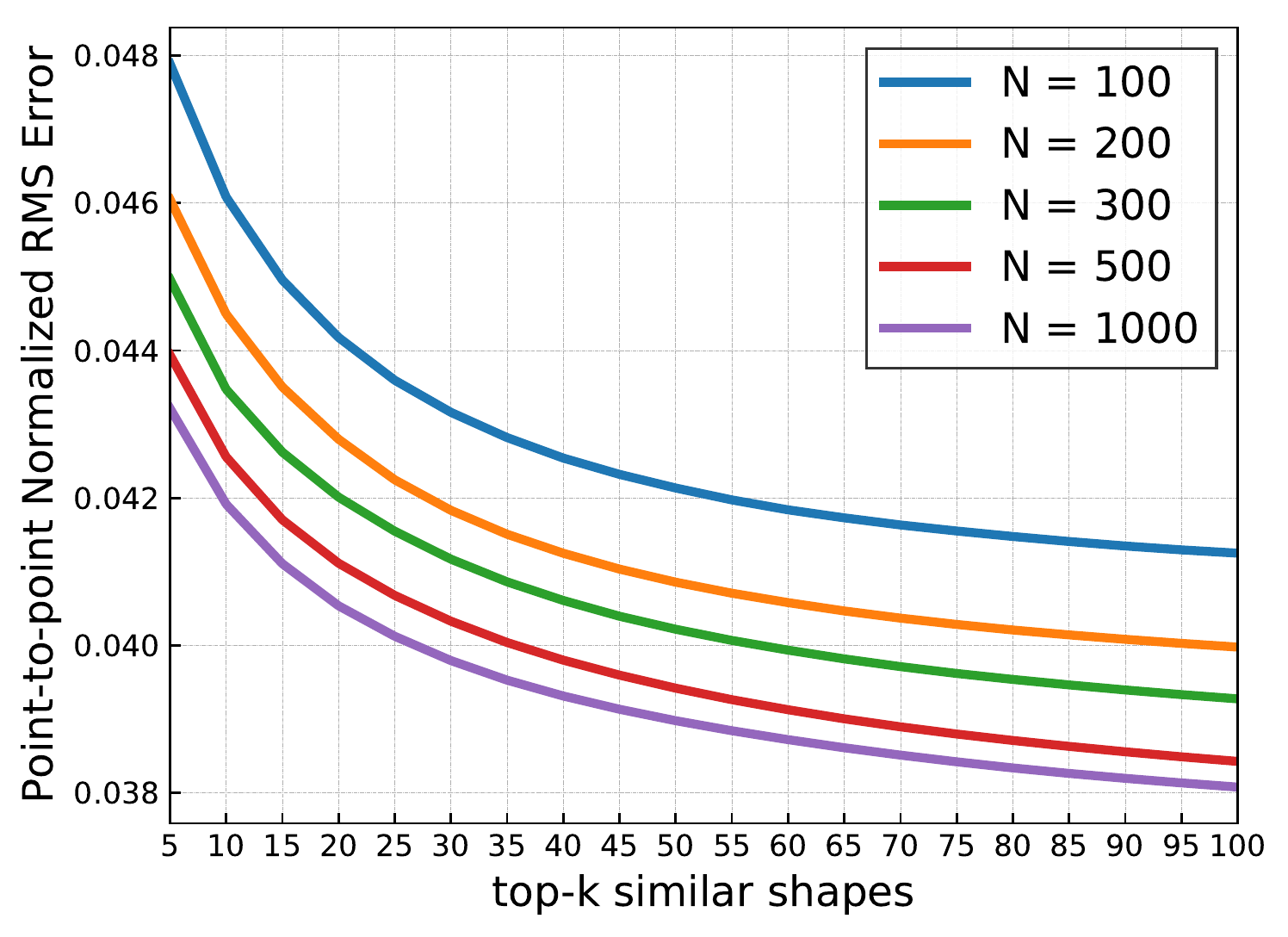}
	\caption{Face shape reconstruction based on $k$ nearest exemplar shapes in dictionary with size $N$. The results are obtained using COFW dataset.}
	\label{fig:dict_N_K}
\end{figure} 

In the face shape dictionary training procedure, the 300-W training set and semifrontal face of the Menpo\cite{zafeiriou2017menpo} dataset are used to train 68-point face shape dictionaries. Additionally, the WFLW training set is used to train 98-point face shape dictionaries. First, affine transformation is performed with the ground truth coordinates of the pupil and the coordinates of the midpoint to make the face canonical. Then, the face shapes are normalized by converting the coordinates of each landmark to a $128\times128$ space. K-means algorithm is utilized to cluster normalized face shapes to reduce spatial redundancy and improve the computational efficiency. As shown in \autoref{fig:dict_N_K}, we tested different dictionary sizes $N$ and different numbers $k$ of face shapes for reconstruction. Finally $N$ and $k$ are set as 500 and 100, respectively. Therefore, the face shapes are reconstructed by 100 most similar shapes in dictionary with size 500. The reconstruction coefficients are computed by the least squares method and ridge regression. The regularization parameter of ridge regression is set to 60. In\autoref{equ:loss_G}, $a$ and $b$ are set to 1 and 0.5, respectively.

\begin{figure*}[!t]
	\centering
	\includegraphics[width=\linewidth]{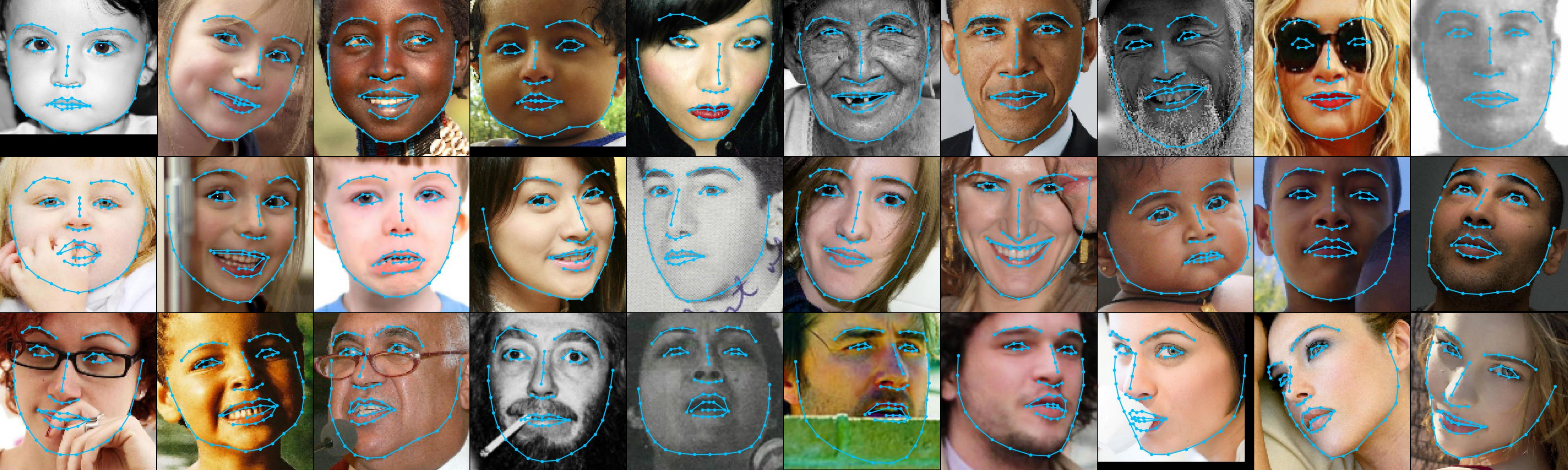}
	\caption{MSM example outputs using 300-W dataset. For clarity of illustration, detected key points are connected to show dotted face shapes.}
	\label{fig:300W}
\end{figure*}

\begin{table}[ht]
	\centering
	\setlength{\abovecaptionskip}{2pt}
	\caption{NRMSE (\%) of face alignment results using 300-W dataset.}
	\label{table:300W_result}
	\centering
	\begin{tabular}{ccccc}
		\toprule[1pt]
		\multirow{2}{*}{Method}	&\multirow{2}{*}{Year}	&Common   &Challenge   &\multirow{2}{*}{Fullset} \\ 
		&												&Subset		&Subset	         &	\\ \midrule[0.5pt]
		\multicolumn{5}{c}{NRMSE (inter-pupils) (\%)}	\\ \midrule[0.5pt] 
		LBF\cite{ren2016face}	&2014	&4.95	&11.98	&6.32	\\
		TCDCN\cite{zhang2014facial}	&2014	&4.80	&8.60	&5.54	\\
		CFSS\cite{zhu2015face}	&2015	&4.73	&9.98	&5.76	\\
		MDM\cite{trigeorgis2016mnemonic} &2016	&4.83	&10.14	&5.88	\\
		RAR\cite{xiao2016robust}	&2016	&4.12	&8.35	&4.94	\\
		DAN\cite{kowalski2017deep}	&2017	&4.42	&7.57	&5.03	\\
		TSR\cite{lv2017deep}	&2017	&4.36	&7.56	&4.99	\\
		SHN\cite{yang2017stacked}	&2017	&4.12	&7.00	&4.68	\\
		LAB\cite{wu2018look}	&2018	&\textbf{3.42}	&6.98	&\textbf{4.12}	\\
		DCFE\cite{valle2018deeply}	&2018	&3.83	&7.54	&4.55	\\
		3DDE\cite{valle2019face}	&2019	&3.73	&7.10	&4.39	\\
		AGCFN\cite{liu2017adaptive}	&2019	&3.73	&7.24	&4.42	\\	\midrule[0.5pt]
		\textbf{MSM}	&2019	& 3.74	&\textbf{6.97}	&4.38	\\	\midrule[0.5pt]
		\multicolumn{5}{c}{NRMSE (inter-ocular) (\%)}	\\ \midrule[0.5pt]
		DAN\cite{kowalski2017deep}	&2017	&3.19	&5.24	&3.59	\\
		PCD-CNN\cite{kumar2018disentangling}	&2018	&3.67	&7.62	&4.44	\\
		SAN\cite{dong2018style}	&2018	&3.34	&6.60	&3.98	\\
		LAB\cite{wu2018look}	&2018	&2.98	&5.19	&3.49	\\
		DCFE\cite{valle2018deeply}	&2018	&2.76	&5.22	&3.24	\\
		ODN\cite{Zhu_2019_CVPR}	&2019 &3.56	&6.67	&4.17	\\
		3DDE\cite{valle2019face}	&2019	&\textbf{2.69}	&4.92	&3.13	\\
		DeCaFA\cite{Dapogny_2019_ICCV}	&2019	&2.93	&5.26	&3.39	\\	\midrule[0.5pt]
		\textbf{MSM}	&2019	&2.70	&\textbf{4.83}	&\textbf{3.11}	\\
		\bottomrule[1pt]
	\end{tabular}
\end{table}

Our model is implemented on Ubuntu 18.04 with a NVIDIA GTX1080 (8GB) GPU and an Intel Core 7500 CPU @3.4 GHz$\times4$. Training the ST-GAN and stacked hourglass network took around 8 hours and 6 hours respectively. The Python implementation process images at 14 FPS on average, the CNN part (the ST-GAN and stacked hourglass network) took around 50 ms and the shape reconstruction took around 20 ms per image.

\subsection{Experiment using 300-W dataset}
Many existing methods have established a series of impressive results on this dataset. In\autoref{table:300W_result}, we compare our results with LBF\cite{ren2016face}, TCDCN\cite{zhang2014facial}, CFSS\cite{zhu2015face}, MDM\cite{trigeorgis2016mnemonic}, RAR\cite{xiao2016robust}, DAN\cite{kowalski2017deep},  TSR\cite{lv2017deep}, SHN\cite{yang2017stacked}, LAB\cite{wu2018look}, DCFE\cite{valle2018deeply}, 3DDE\cite{valle2019face}, PCD-CNN\cite{kumar2018disentangling}, SAN\cite{dong2018style}, DeCaFA\cite{Dapogny_2019_ICCV}, AGCFN\cite{liu2019attention} and ODN\cite{Zhu_2019_CVPR} are also used in\autoref{table:300W_result}.

First, we report the NRMSE results on 300-W dataset of the proposed MSM method and those of other methods in\autoref{table:300W_result}. For the Challenge Subset of 300-W, the MSM achieves an inter-pupils NRMSE of $6.97\%$ and an inter-ocular NRMSE of $4.83\%$. This demonstrates the MSM is robust to handling face under difficult scenarios such as large pose, lighting and occlusion, etc. For the Common Subset and Fullset of 300-W, the inter-pupils NRMSE values of LAB is slightly better than those of the MSM. However, the LAB is much more computational expensive due to a network architecture using eight stacked hourglass modules versus two stacked hourglass modules in the MSM. For the Common Subset and Fullset of 300-W, comparable inter-ocular NRMSE values are obtained by the 3DDE using a UNet-based network and MSM using two stacked hourglass modules in which MSM obtained slightly higher and slightly lower NRMSE values respectively in the Common Subset and Fullset.\autoref{fig:300W} shows the MSM results using 300-W dataset.

\begin{table}[!t]
	\centering
	\setlength{\abovecaptionskip}{2pt}
	\caption{Inter-ocular NRMSE (\%), failure rate (\%) and AUC of face alignment results using 300-W private test set.}
	\label{table:300w_private_result}
	\centering
	\begin{tabular}{cccc}	\toprule[1pt]
		Method							&NRMSE (\%)		&Failure (\%)	&AUC	\\	\midrule[0.5pt]	
		CFSS\cite{zhu2015face}			&-			&12.30			&0.4132	\\
		MDM\cite{trigeorgis2016mnemonic}&5.05		&6.80			&0.4532	\\
		DAN\cite{kowalski2017deep}		&4.30		&2.67			&0.4700	\\
		SHN\cite{yang2017stacked}		&4.05		&-				&-		\\
		DCFE\cite{valle2018deeply}		&3.88		&1.83			&0.5242	\\
		AGCFN\cite{liu2019attention}		&3.82		&1.60			&0.5252	\\	\midrule[0.5pt]	
		\textbf{MSM}								&\textbf{3.81}	&\textbf{1.50}  &\textbf{0.5262}	\\
		\bottomrule[1pt]
	\end{tabular}
\end{table}

\begin{figure}[!t]
	\centering
	\includegraphics[width=0.6\linewidth]{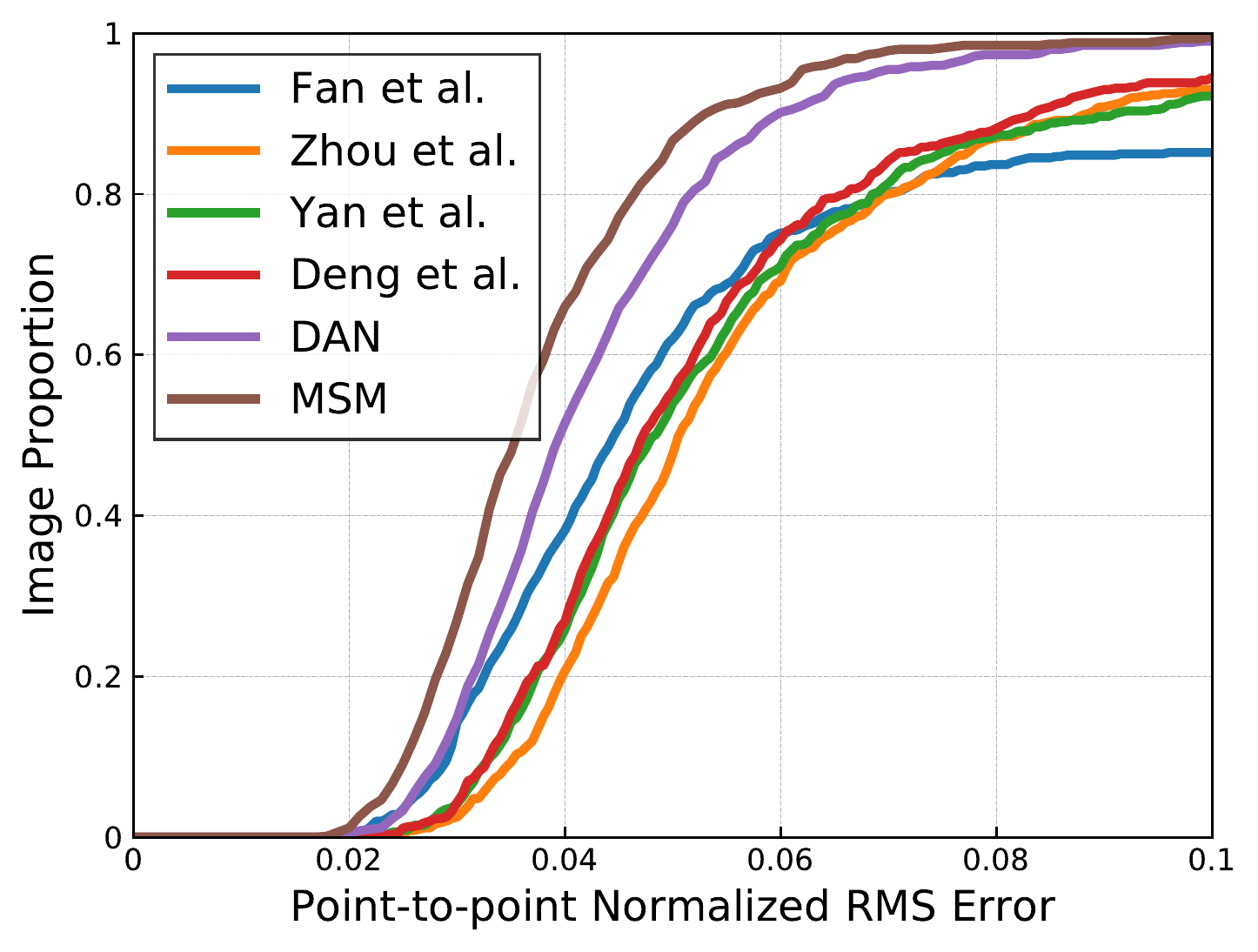}
	\caption{CED curves of face alignment results using 300-W private test set. }
	\label{fig:ced of 300W}
\end{figure}

For the 300-W private test set, the comparison of NRMSE, failure rate and AUC are shown in\autoref{table:300w_private_result} indicate that the MSM outperforms all other methods in NRMSE values, failure rate and AUC with the exception that the DCFE achieved an AUC of 0.5242 versus the MSM of 0.5262.

\begin{figure*}[!t]
	\centering
	\includegraphics[width=\linewidth]{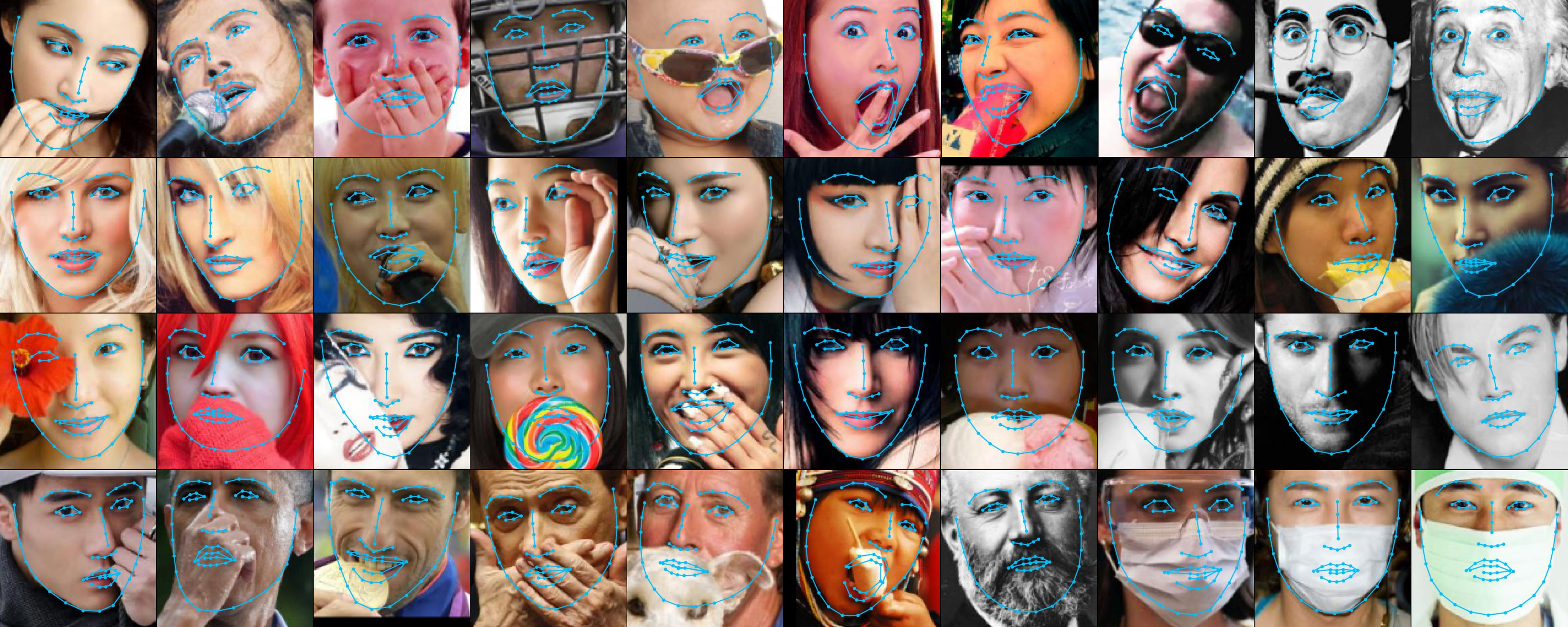}
	\caption{MSM example outputs using COFW dataset subject to various occlusion, such as hands, glasses, food, and mask covering a wide range of faces.}
	\label{fig:COFW}
\end{figure*}

We compare the CED curves obtained by the DAN, the method proposed by Fan \textit{et al.}\cite{fan2016approaching}, Zhou \textit{et al.}\cite{zhou2013extensive}, Yan \textit{et al.}\cite{yan2013learn} and Deng \textit{et al.}\cite{deng2016m3}. As shown in\autoref{fig:ced of 300W}, MSM obtained the lowest point-to-point NRMSE values as compared to other methods.

Although 300-W is the most widely used face alignment dataset, its small sample size and relatively simple face images limit its scope to be used for comprehensive evaluation on the performance of an algorithm under a broad range of conditions.

\begin{table}[!t]
	\centering
	\setlength{\abovecaptionskip}{2pt}
	\caption{NRMSE (\%) and failure rate (\%) of face alignment results using COFW dataset.}
	\label{table:cofw_result}
	\centering
	\begin{tabular}{C{1.51cm}C{1.71cm}C{1.3cm}C{0.75cm}C{0.75cm}}	\toprule[1pt]
		\multirow{3}{*}{Method}			&\multirow{2}{*}{Training}	&inter-pupils	&\multicolumn{2}{c}{inter-ocular}	\\
		&\multirow{2}{*}{Set}			&NRMSE (\%)			&NRMSE (\%)		&Failure (\%)	\\	\midrule[0.5pt]	
		RCPR\cite{burgos2013robust}		&300-W		&12.27	 		&8.76	 	&20.12	\\
		TCDCN\cite{zhang2014facial}		&300-W			&10.72	&7.66	 &16.17	\\
		HPM\cite{ghiasi2014occlusion}	&300-W			&9.40	&6.72	&6.71	\\
		CFSS\cite{zhu2015face}			&300-W			&8.80	&6.28	&9.07	\\
		SHN\cite{yang2017stacked}		&300-W, Menpo			&5.60	&4.00	&\textbf{0} \\
		JMFA\cite{deng2019joint}		&300-W, Menpo			&5.58	&-		&-	\\
		LAB\cite{wu2018look}			&300-W			&- 	&4.62	 &2.17 	\\
		ODN\cite{Zhu_2019_CVPR}			&300-W &-		&5.30	&- \\
		\midrule[0.5pt]
		\textbf{MSM}					&300-W			&\textbf{5.50}	&\textbf{3.90}  &\textbf{0}	\\
		\bottomrule[1pt]
	\end{tabular}
\end{table}

\begin{figure}[ht]
	\centering
	\includegraphics[width=0.6\linewidth]{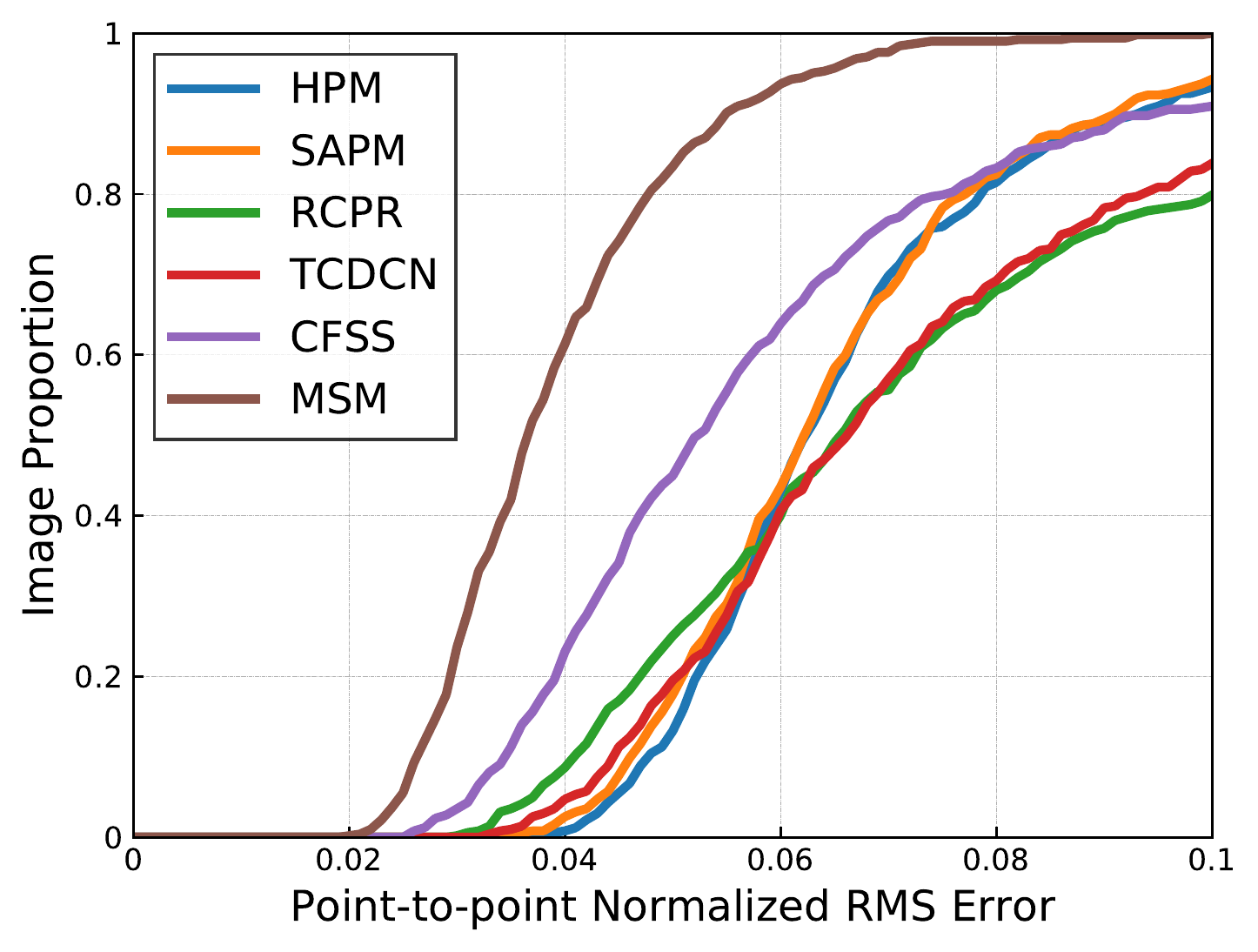}
	\caption{CED curves of face alignment results using COFW dataset.}
	\label{fig:ced of COFW}
\end{figure}

\subsection{Experiment using COFW dataset}

To evaluate the robustness to occlusion of the MSM method subject to various occluded face images, the COFW dataset is used which is regarded as a challenging dataset for existing state-of-the-art face alignment methods. In\autoref{table:cofw_result}, various methods including RCPR, TCDCN, HPM\cite{ghiasi2014occlusion}, CFSS, SHN, JMFA\cite{deng2019joint}, AGCFN and LAB are compared. The MSM was trained on the 300-W dataset with a total of 3148 face training images. As shown in\autoref{table:cofw_result}, the MSM achieved the lowest inter-pupils NRMSE of 5.50\% and the lowest inter-ocular NRMSE value of 3.90\% with failure rate of 0\%. These reflect the effectiveness of MSM in managing faces under heavy occlusion. The NRMSE values for SHN and JMFA are slightly higher than those of the MSM method. It should be noted that the training sets of both the SHN and the JMFA are much larger than that of the MSM in which the SHN and the JMFA include the 300-W and Menpo\cite{zafeiriou2017menpo} training sets, for a total of 9360 face images, which is almost three times more images than that of the MSM.

\autoref{fig:ced of COFW} shows the CED curves which indicate the MSM outperforms other methods (including SAPM\cite{ghiasi2015using}) by a large margin on the COFW dataset. Example results obtained from COFW are given in\autoref{fig:COFW}.

\begin{table*}[!t]
	\centering
	\setlength{\abovecaptionskip}{2pt}
	\caption{NRMSE (\%), failure rate (\%) and AUC of face alignment results using WFLW dataset.}
	\label{table:wflw_result}
	\centering
	\begin{tabular}{ccC{1.1cm}C{1.1cm}C{1.1cm}C{1.1cm}C{1.1cm}C{1.1cm}C{1.1cm}}
		\toprule[1pt]
		Metric	&Method	&Fullset &Pose   &Expression	&Illumination	&Make-up	&Occlusion	&Blur	\\	\midrule[0.5pt]
		\multirow{6}{*}{NRMSE (\%)} 
		&SDM\cite{xiong2013supervised}	&10.29	&24.10	&11.45	&9.32	&9.38	&13.03	&11.28	\\
		&CFSS\cite{zhu2015face}	&9.07	&21.36	&10.09	&8.30	&8.74	&11.76	&9.96	\\
		&DVLN\cite{wu2017leveraging}	&6.08	&11.54	&6.78	&5.73	&5.98	&7.33	&6.88	\\
		&LAB\cite{wu2018look}	&5.27	&10.24	&5.51	&5.23	&5.15	&6.79	&6.32	\\
		&3DDE\cite{valle2019face}	&4.68	&8.62	&5.21	&4.65	&4.60	&5.77	&5.41	\\
		&DeCaFA\cite{Dapogny_2019_ICCV}	&4.62	&8.11	&\textbf{4.65}	&\textbf{4.41}	&4.63	&\textbf{5.74}	&5.38	\\	\midrule[0.5pt]
		&\textbf{MSM}	&\textbf{4.60}	&\textbf{8.01}	&4.81	&4.58	&\textbf{4.47}	&5.85	&\textbf{5.28}	\\ \midrule[0.5pt]
		\multirow{6}{*}{Failure (\%)} 
		&SDM\cite{xiong2013supervised}	&29.40	&84.36	&33.44	&26.22	&27.67	&41.85	&35.32	\\
		&CFSS\cite{zhu2015face}	&20.56	&66.26	&23.25	&17.34	&21.84	&32.88	&23.67		\\
		&DVLN\cite{wu2017leveraging}	&10.84	&46.93	&11.15	&7.31	&11.65	&16.30	&13.71	\\
		&LAB\cite{wu2018look}	&7.56	&28.83	&6.37	&6.73	&7.77	&13.72	&10.74	\\
		&3DDE\cite{valle2019face}	&5.04	&22.39	&5.41	&3.86	&6.79	&9.37	&6.72	\\
		&DeCaFA\cite{Dapogny_2019_ICCV}	&4.84	&21.4	&3.73	&\textbf{3.22}	&6.15	&\textbf{9.26}	&6.61	\\	\midrule[0.5pt]
		&\textbf{MSM}	&\textbf{4.28}	&\textbf{16.87}	&\textbf{2.87}	&3.72	&\textbf{4.37}	&9.36	&\textbf{5.95}	\\ \midrule[0.5pt]
		\multirow{6}{*}{AUC} 
		&SDM\cite{xiong2013supervised}	&0.3002	&0.0226	&0.2293	&0.3237	&0.3125	&0.2060	&0.2398	\\
		&CFSS\cite{zhu2015face}	&0.3659	&0.0632	&0.3157	&0.3854	&0.3691	&0.2688	&0.3037	\\
		&DVLN\cite{wu2017leveraging}	&0.4551	&0.1474	&0.3889	&0.4743	&0.4494	&0.3794	&0.3973	\\
		&LAB\cite{wu2018look}	&0.5323	&0.2345	&0.4951	&0.5433	&0.5394	&0.4490	&0.4630	\\
		&3DDE\cite{valle2019face}	&0.5544	&0.2640	&0.5175	&0.5602	&0.5536	&0.4692	&0.4957	\\	
		&DeCaFA\cite{Dapogny_2019_ICCV}	&0.5630	&0.2920	&0.5460	&\textbf{0.5790}	&\textbf{0.5750}	&\textbf{0.4850}	&0.4940	\\\midrule[0.5pt]
		&\textbf{MSM}	&\textbf{0.5671}	&\textbf{0.3091}	&\textbf{0.5478}	&0.5725	&0.5711	&0.4849	&\textbf{0.5073}	\\
		\bottomrule[1pt]
	\end{tabular}
\end{table*}

\begin{figure*}[!t]
	\centering
	\includegraphics[width=\linewidth]{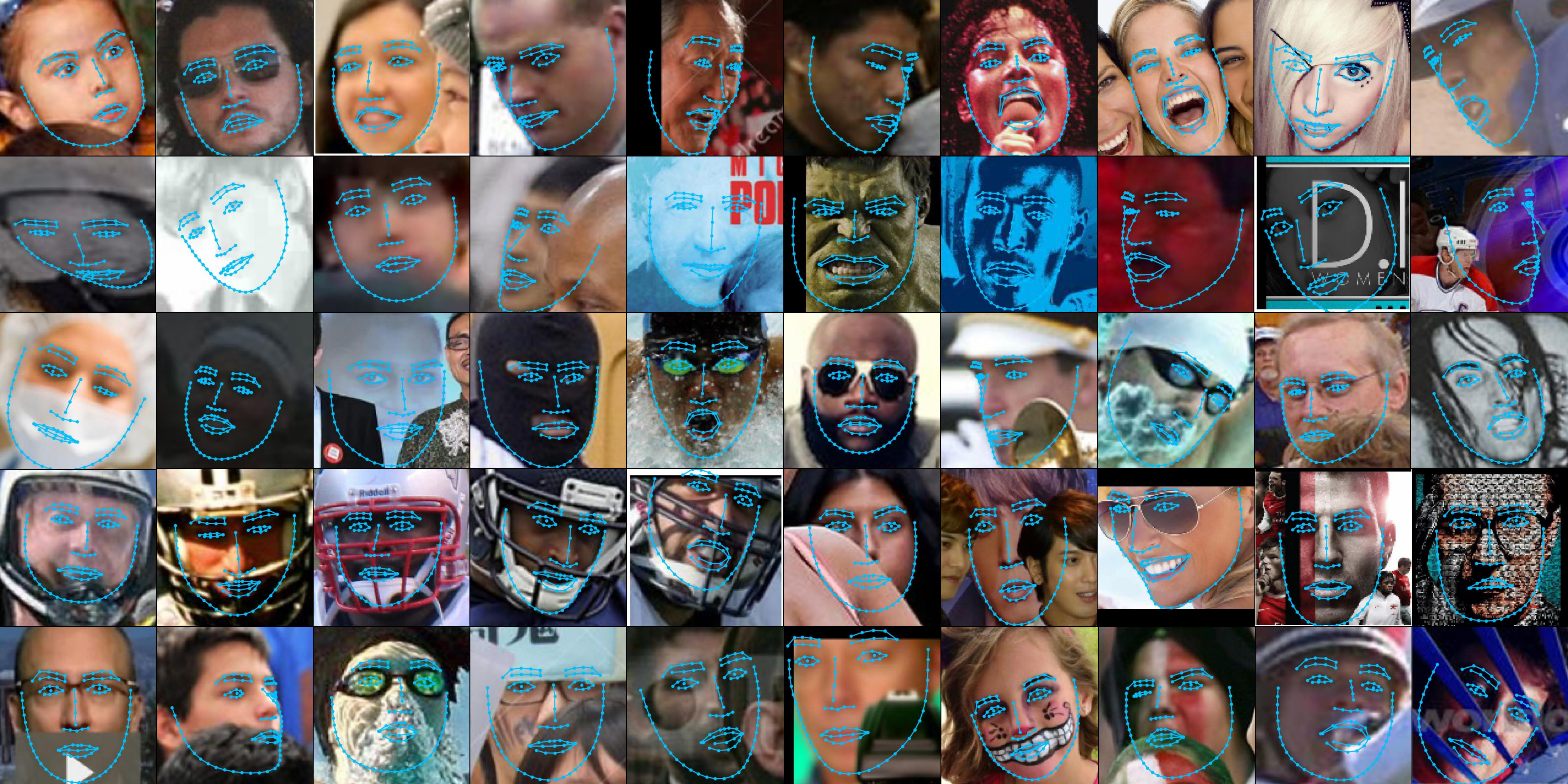}
	\caption{MSM example outputs using WFLW dataset subject to extremely challenge cases, such as illumination, large pose, occlusion and disturbing background, etc.}
	\label{fig:WFLW}
\end{figure*}

\subsection{Experiment using WFLW dataset}

\begin{table*}[ht]
	\centering
	\setlength{\abovecaptionskip}{2pt}
	\caption{Comparison of NRMSE (\%) using WFLW dataset with different configurations.}
	\label{table:ablation_wflw}
	\centering
	\begin{tabular}{cC{1.1cm}C{1.1cm}C{1.1cm}C{1.1cm}C{1.1cm}C{1.1cm}C{1.1cm}}
		\toprule[1pt]
		Method			&Fullset	&Pose   &Expression	&Illumination	&Make-up	&Occlusion	&Blur	\\	\midrule[0.5pt]
		Res-50			&5.73		&11.28	&6.13		&5.65			&5.80		&6.98		&6.51	\\
		ST-GAN + Res-50	&5.42		&10.65	&6.00		&5.31			&5.39		&6.57		&6.23	\\
		HG				&5.41		&10.03	&5.56		&5.54			&6.03		&7.00		&6.25	\\
		ST-GAN + HG		&4.81		&8.49	&5.09		&4.75			&4.70		&6.16		&5.51	\\
		HG + SR		&5.17		&9.49	&5.42		&5.38			&5.74		&6.60		&6.08	\\
		ST-GAN + HG + SR	&4.60		&8.01	&4.81		&4.58			&4.47		&5.85		&5.28	\\
		\bottomrule[1pt]
	\end{tabular}
\end{table*}

\begin{table}[!t]
	\centering
	\setlength{\abovecaptionskip}{2pt}
	\caption{Comparisons of NRMSE (\%) and failure rate (\%) using COFW dataset with different configurations.}
	\label{table:ablation_cofw}
	\centering
	\begin{tabular}{ccc}\toprule[1pt]
		Method	 		 &NRMSE (\%)		&Failure (\%)	\\	\midrule[0.5pt]
		Res-50			 &4.76		&4.54\\
		ST-GAN + Res-50 	 &4.23		&3.81\\
		HG				 &4.64		&6.52\\
		ST-GAN + HG		 &4.34		&5.23\\
		HG + SR		 	 &4.10		&0.99\\
		ST-GAN + HG + SR	 &3.95		&0.99\\
		\bottomrule[1pt]
	\end{tabular}
\end{table}

\begin{figure}[ht!]
	\centering
	\includegraphics[width=0.6\linewidth]{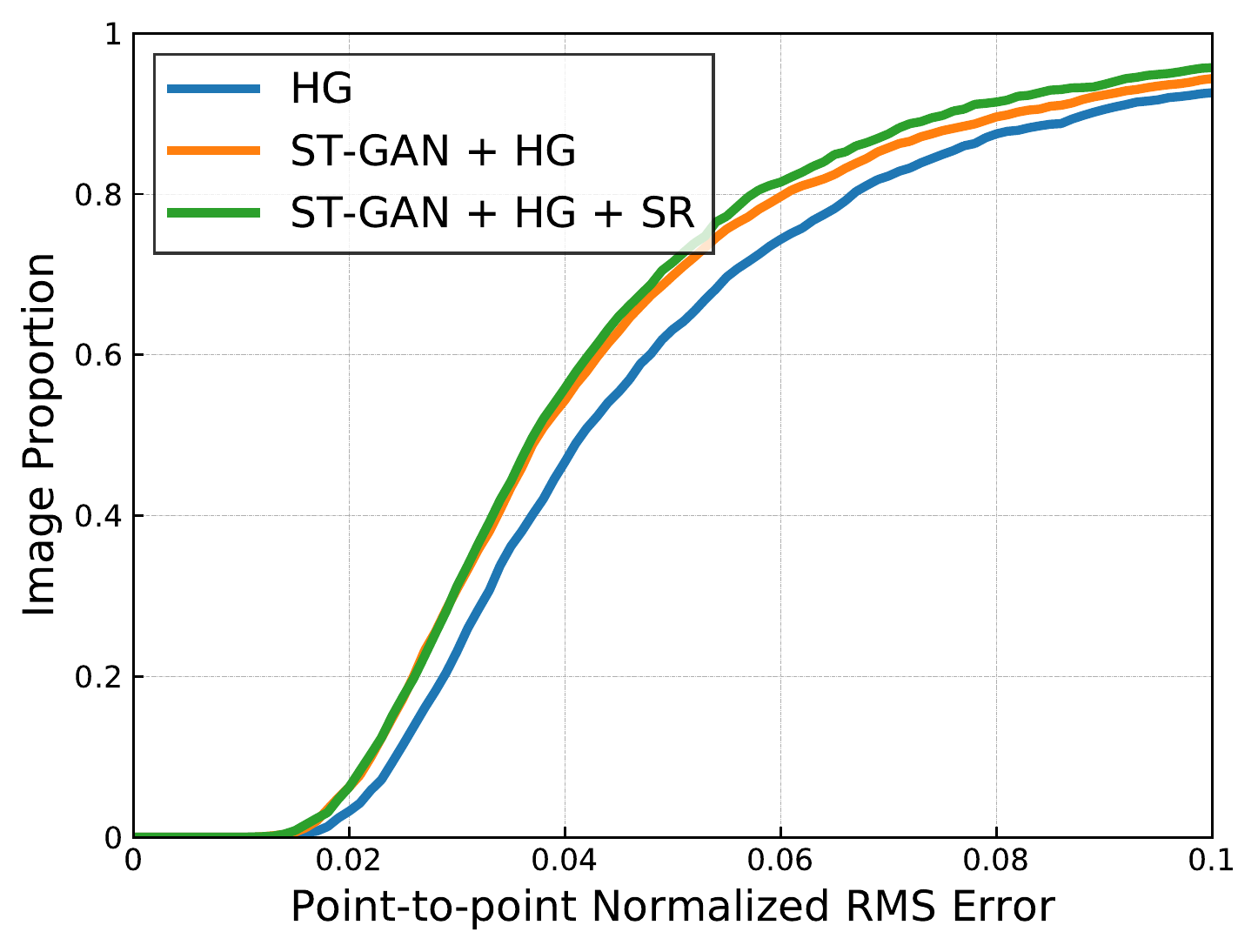}
	\caption{Comparisons of CED curves using WFLW dataset with different configurations.}
	\label{fig:ced ablation_on_wflw}
\end{figure}

The landmark configurations of this dataset is different from above datasets, all images in WFLW dataset are annotated by 98-points manually. For comprehensive analysis of existing state-of-the-art methods, the dataset contains various type of challenge including large pose, illumination, blur, occlusion and excessive disturbing background, etc. Since WFLW is a newly released dataset, we compare the proposed method with a number of methods including ESR, SDM, CFSS, DVLN\cite{wu2017leveraging}, LAB, 3DDE and DeCaFA\cite{Dapogny_2019_ICCV}. We report the NRMSE (inter-ocular), failure rate and AUC on the test set and six subsets of WFLW. As shown in\autoref{table:wflw_result}, the MSM method outperforms all other state-of-the-art methods in terms of the NRMSE, failure rate and AUC. An exception is for the case of an NRMSE value of 5.77\% (occlusion subset) obtained by the 3DDE versus 5.85\% obtained by the MSM. Note that the input images of 3DDE are cropped by ground-truth bounding box, which is much more beneficial to landmark localisation task. However, MSM still outperforms 3DDE using the provided bounding box in all other metrics. The MSM results using WFLW dataset are shown in\autoref{fig:WFLW}.

\subsection{Experimental results on ablation study}

In this subsection the proposed method is evaluated by different configurations. The framework consists of several pivotal components including ST-GAN, stacked hourglass network and examplar-based face shape reconstruction. Their effectiveness are validated within the framework based on the COFW and WFLW datasets. To further evaluate the robustness of ST-GAN, a 50-layer residual network (Res-50) is introduced to verify whether the ST-GAN is effective to coordinate regression-based method. Since Res-50 requires input images size of $224\times224$, the size of the average pooling kernel in Res-50 is resized from 7 to 4, and the size of the network input is $128\times128$. The results of all ablation experiments use the inter-ocular distance as normalizing factor. Each proposed component was analyzed, i.e., with ST-GAN (labeled as “ST-GAN”), hourglass network (labeled as “HG”), and shape reconstruction (labeled as “SR”), by comparing their NRMSE and failure rates. Note that our baseline is HG, and ST-GAN+HG+SR represents the full MSM method.

\begin{figure*}[!h]
	\centering
	\includegraphics[width=\linewidth]{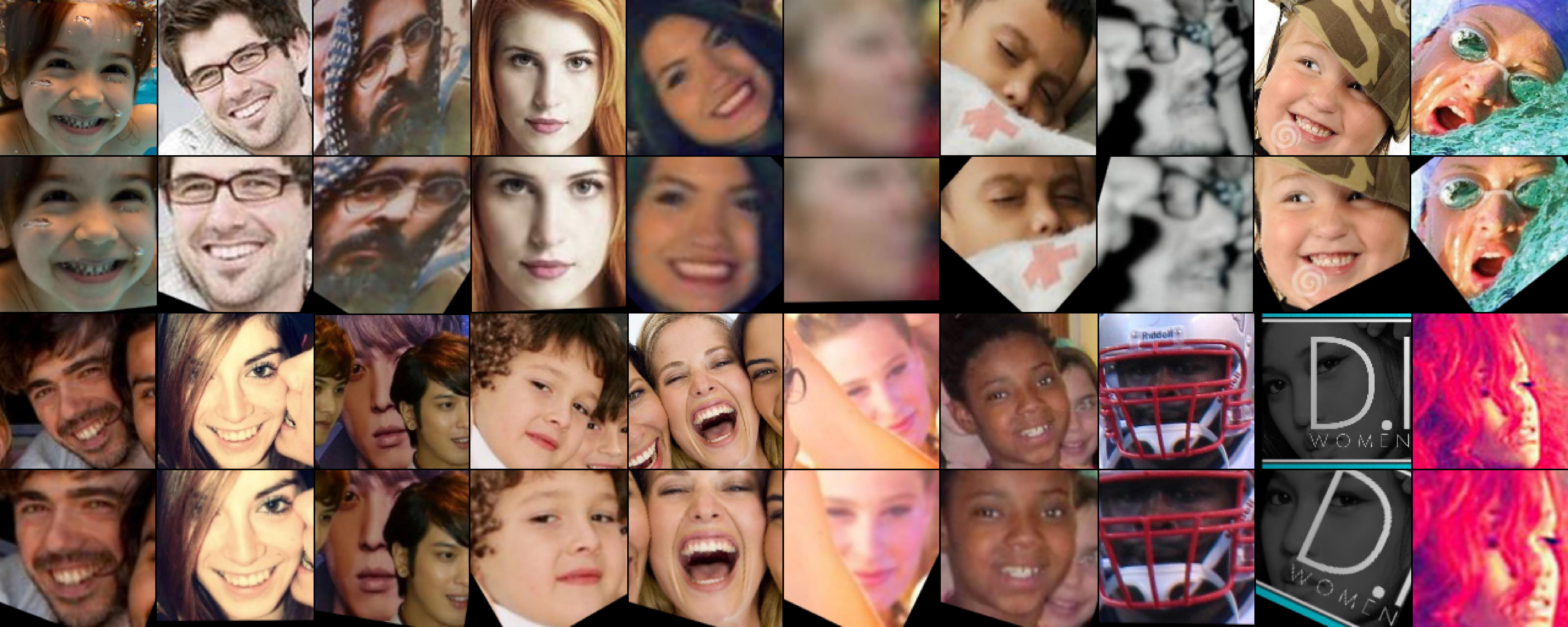}
	\caption{ST-GAN examples outputs using WFLW dataset. Images in first and third rows are cropped by provided bounding boxes. Images in second and fourth rows are obtained by ST-GAN. Note that ST-GAN not only normalizes face but also removes disturbing background areas.}
	\label{fig:stn_results}
\end{figure*}

\begin{figure}[ht!]
	\centering
	\includegraphics[width=0.9\linewidth]{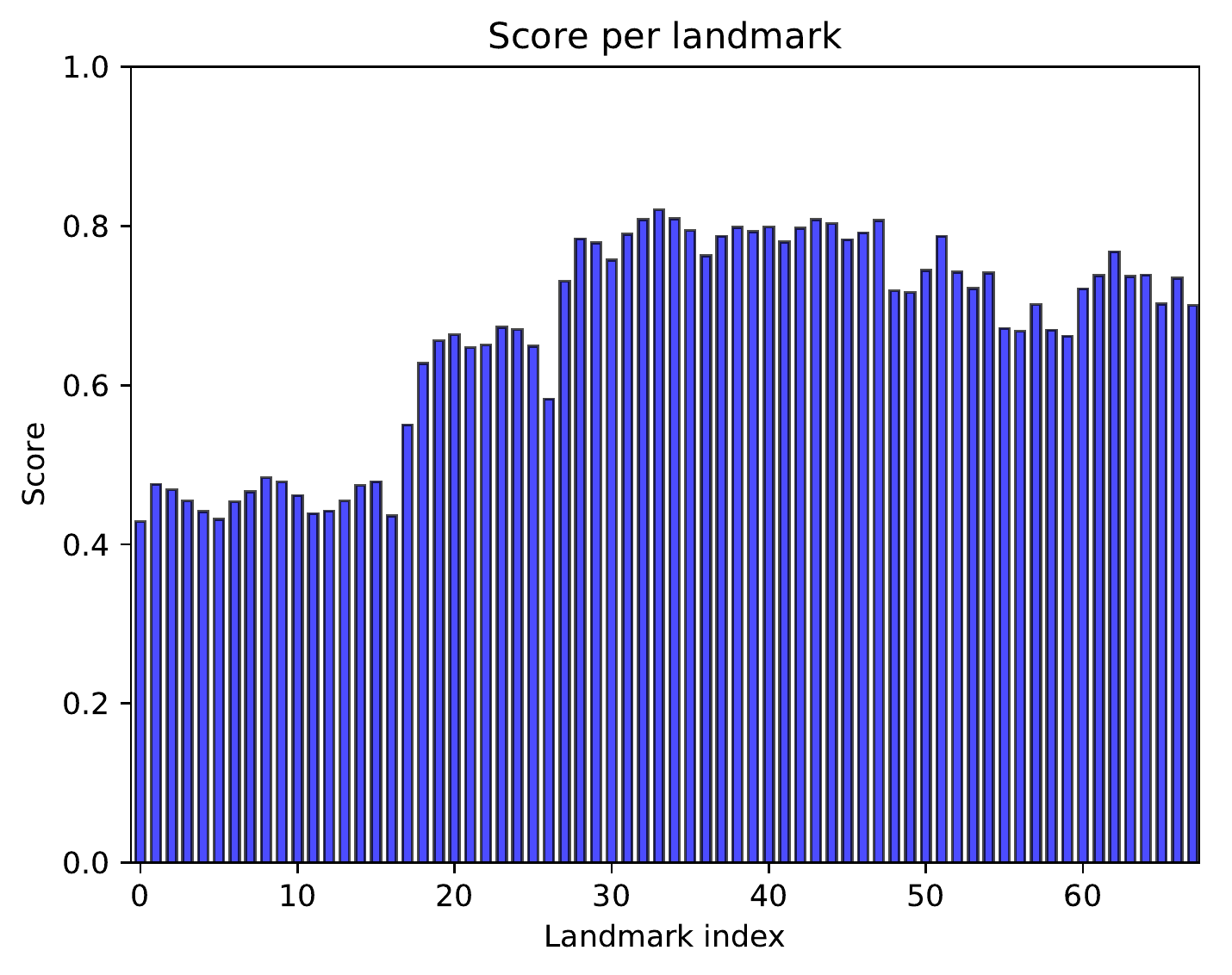}
	\caption{Score distribution related to each landmark using COFW dataset.}
	\label{fig:score_per_landmark_on_cofw}
\end{figure}

\begin{figure}[ht!]
	\centering
	\includegraphics[width=0.6\linewidth]{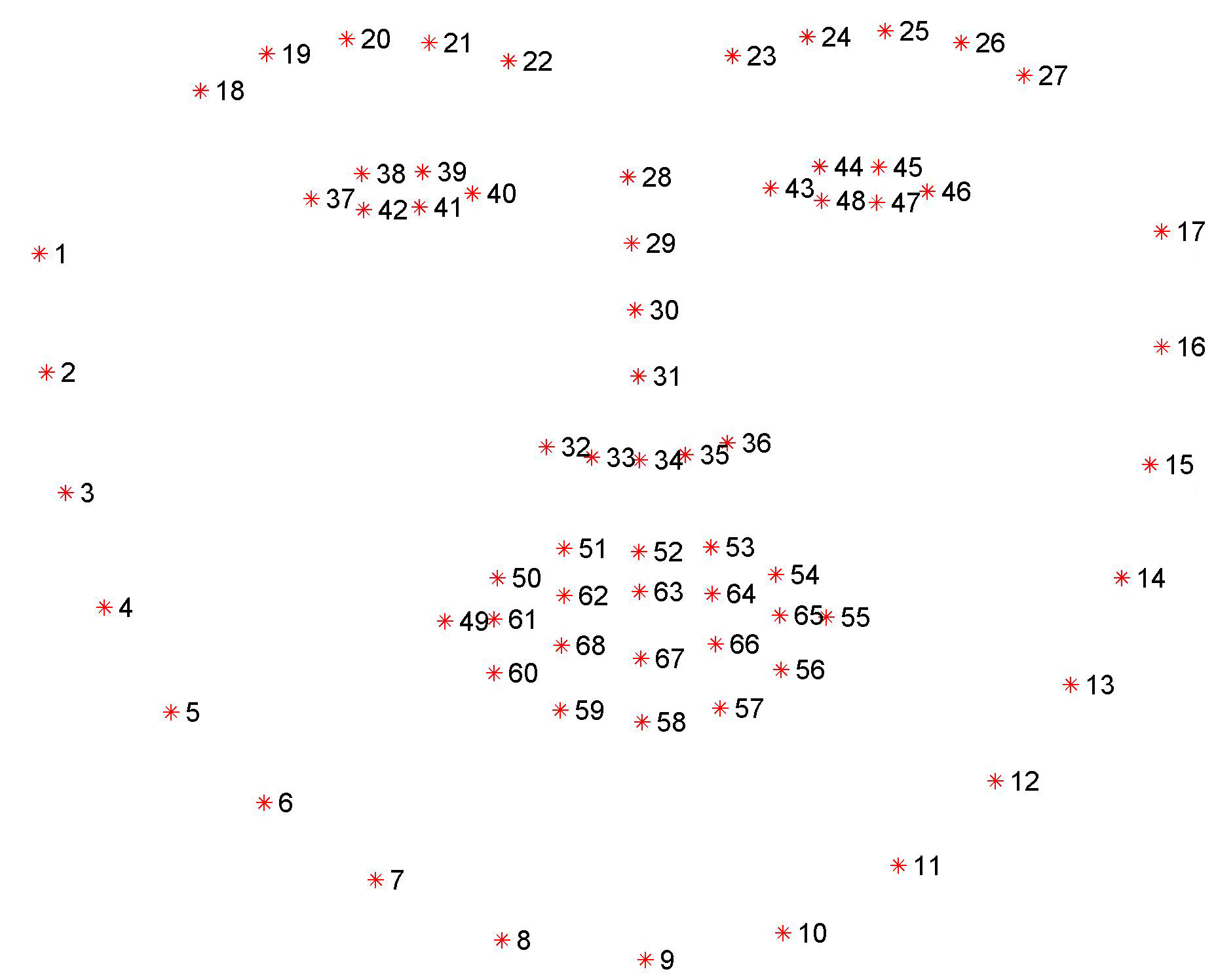}
	\caption{Landmark definition of the 68-point datasets including 300-W and COFW.}
	\label{fig:landmark definition}
\end{figure}

\autoref{table:ablation_wflw} and\autoref{table:ablation_cofw} show the NRMSE values and failure rates obtained by different configurations of our framework evaluated on the COFW and WFLW datasets. When combined with the ST-GAN, the Res-50 network reduces the NRMSE from 4.76\% to 4.23\%, and the hourglass network decrease the NRMSE from 4.64\% to 4.34\%. This result demonstrates that the proposed ST-GAN method improved the performance of the face alignment task because STN can remove the translation, scale and rotation variation in each face, which can further reduce the variance in the regression target. Note that our method can effectively normalize face images to canonical poses and simultaneously remove unnecessary background. Compared with the baseline (HG) of our work, the innovations introduced in this paper exhibit a certain improvement for each subset of the WFLW dataset. These results demonstrate that in various difficult situations, the scoring scheme and face shape reconstruction method can be used to accurately locate difficult key points, not just in the case of occlusion. In\autoref{fig:ced ablation_on_wflw}, the CED curves show that ST-GAN+HG+SR which representing the full MSM method outperforms the other two configurations. Examples of the outputs obtained by the proposed ST-GAN on the WFLW dataset are shown in\autoref{fig:stn_results}.

\begin{table}[!t]
	\centering
	\setlength{\abovecaptionskip}{2pt}
	\caption{Comparison of different configurations of threshold $K$ using COFW dataset, ``contour" denotes the threshold $K$ of the landmarks at contour, ``contour" denotes the threshold $K$ of the landmarks at contour, ``facial features" denotes the threshold $K$ of the landmarks at facial features}
	\label{table:ablation_threshold_K}
	\centering
	\begin{tabular}{ccc}\toprule[1pt]
		\multicolumn{2}{c}{Configuration}	 		 &\multirow{2}{*}{NRMSE (\%)}	\\	
		contour		&facial features						 \\ \midrule[0.5pt]
		0.3			&0.4	&5.59			\\
		0.3			&0.5	&5.57			\\
		0.3			&0.6	&5.58			\\
		0.3			&0.7	&5.65			\\					
		0.4			&0.4	&5.55			\\
		0.4			&0.5	&5.54			\\
		0.4			&0.6	&5.50			\\
		0.4			&0.7	&5.61			\\		
		0.5			&0.4	&5.62			\\
		0.5			&0.5	&5.60			\\
		0.5			&0.6	&5.63			\\
		0.5			&0.7	&5.66			\\	\bottomrule[1pt]
	\end{tabular}
\end{table}

Finally, we discuss the setting of the threshold $K$ for distinguishing the reliability of landmarks. To this end, we performed a statistical analysis of the scores for each landmark of each sample on the COFW dataset, as shown in\autoref{fig:score_per_landmark_on_cofw}. As can be seen from the definition of the landmarks in\autoref{fig:landmark definition}, landmarks 1 to 17 in the contour of the face obtain significantly lower scores. This is because the features of the face contours are relatively simple. Conversely, features near the facial features are significantly more discriminative, thus landmarks at these locations have higher scores. From the above analysis, we can draw a conclusion that it is unreliable to set the same threshold $K$ for all landmarks to distinguish the localization quality. The landmarks at the contour of the face should be set with lower thresholds, while the landmarks at the facial features of the face are in contrast. Therefore, we verified several different threshold configurations, as shown in\autoref{table:ablation_threshold_K}. Finally, the setting for the threshold $K$ is: landmarks at the contour is 0.4, and landmarks at the facial features is 0.6.

\section{Conclusion}
In this paper a multistage model has been presented for robust face alignment. Our method leverages the best advantages of STNs, CNNs and exemplar-based shape constraints. Benefiting from the robust spatial transformation of the ST-GAN, the input image is warped to an alignment-friendly state. The stacked hourglass network provides accurate localization to landmarks that contain rich local information. The intensity of the heatmap is introduced to distinguish the aligned landmarks from missing landmarks, and the weight of each aligned landmark is determined simultaneously. Finally, with the help of these aligned landmarks, misaligned landmarks is refined by sparse shape constraints. A compact face shape dictionary learned by the K-means algorithm is used to improve the computational efficiency. Extensive experiments and ablation study have been conducted using challenging datasets (300-W, COFW and WFLW), the experimental results and analysis have demonstrated the effectiveness of the proposed multistage model as compared to other state-of-the-art methods. For portable and real-time aplications, multiplierless neural networks\cite{kwan2002multiplierless,256024,1015735,229903,156266,191867,252417,tang1993parameter} can be designed using back propagation\cite{tang1993parameter} and other algorithms for implementing the multistage model. 

Demos are posted on the website at http://101.37.150.44:8088/msm.aspx.



%



\section*{Acknowledgment}
This work was supported in part by the National Natural Science Foundation of China under Grant 61372137, in part by the Natural Science Foundation of Anhui Province under Grant 1908085MF209 and Grant 1708085MF151, and in part by the Natural Science Foundation for the Higher Education Institutions of Anhui Province under Grant KJ2019A0036.

\ifCLASSOPTIONcaptionsoff
  \newpage
\fi



\bibliographystyle{IEEEtran}
\bibliography{TMM}

\end{document}